\documentclass{ecai} 

\usepackage{amsmath,amssymb,mathabx,amsthm} 
\usepackage{latexsym}
\usepackage{enumitem}
\usepackage{array}
\usepackage{subfig}
\usepackage{textcomp}
\usepackage{stfloats}
\usepackage{hyperref}
\usepackage{url}
\usepackage{verbatim}
\usepackage{graphicx}
\usepackage[nocompress]{cite}

\usepackage{xspace}
\usepackage{rotating}
\usepackage{multirow}
\usepackage{booktabs}
\usepackage{colortbl}
\usepackage[table]{xcolor}
\usepackage{bm}
\usepackage[utf8]{inputenc}
\usepackage{setspace}
\usepackage{caption}

\usepackage[linesnumbered,boxed,ruled,vlined]{algorithm2e} 

\SetCommentSty{mycommfont}
\SetKwInput{KwData}{Input}
\SetKwInput{KwResult}{Output}

\hypersetup{
    colorlinks,
    linkcolor={red!50!black},
    citecolor={blue!50!black},
    urlcolor={blue!80!black}
}
\hypersetup{pdfauthor={Name}}

\definecolor{myblue}{RGB}{54,100,139}
\definecolor{myblue2}{RGB}{70,131,180}
\definecolor{mygray}{gray}{0.55}
\definecolor{myg}{gray}{0.85}
\definecolor{mylblue}{RGB}{210, 235, 255}

\newcommand{\HBic}{\mbox{\texttt{HBIC}}\xspace}
\newcommand{\CHUL}{\mbox{\texttt{CHUL}}\xspace}

\newcommand{\BibTeX}{B\kern-.05em{\sc i\kern-.025em b}\kern-.08em\TeX}

\newcolumntype{h}{>{\columncolor{myg}}c}
\newcolumntype{s}{>{\columncolor{mylblue}}c}
\newcolumntype{g}{>{\columncolor{myg}}l}
\graphicspath{{figures/}}

\begin{document}


\begin{frontmatter}

\paperid{123}

\title{HBIC: A Biclustering Algorithm for Heterogeneous Datasets}

\author[A,B]{\fnms{Adán}~\snm{José-García}\thanks{Corresponding Author. Email: adan.josegarcia@univ-lille.fr.}}
\author[A]{\fnms{Julie}~\snm{Jacques}}
\author[B]{\fnms{Clément}~\snm{Chauvet}} 
\author[B,C]{\fnms{Vincent}~\snm{Sobanski}}
\author[A]{\fnms{Clarisse}~\snm{Dhaenens}} 

\address[A]{Univ. Lille, CNRS, Centrale Lille, UMR 9189 CRIStAL, F-59000 Lille, France}
\address[B]{Univ. Lille, Inserm, CHU Lille, U1286 - INFINITE - Institute for Translational Research in Inflammation, F-59000 Lille, France}
\address[C]{Institut Universitaire de France (IUF), Paris, France}

\begin{abstract}

Biclustering is an unsupervised machine-learning approach aiming to cluster rows and columns simultaneously in a data matrix. Several biclustering algorithms have been proposed for handling numeric datasets. However, real-world data mining problems often involve heterogeneous datasets with mixed attributes. To address this challenge, we introduce a biclustering approach called \HBic, capable of discovering meaningful biclusters in complex heterogeneous data, including numeric, binary, and categorical data. The approach comprises two stages: bicluster generation and bicluster model selection. In the initial stage, several candidate biclusters are generated iteratively by adding and removing rows and columns based on the frequency of values in the original matrix. In the second stage, we introduce two approaches for selecting the most suitable biclusters by considering their size and homogeneity. Through a series of experiments, we investigated the suitability of our approach on a synthetic benchmark and in a biomedical application involving clinical data of systemic sclerosis patients. The evaluation comparing our method to existing approaches demonstrates its ability to discover high-quality biclusters from heterogeneous data. Our biclustering approach is a starting point for heterogeneous bicluster discovery, leading to a better understanding of complex underlying data structures.

\end{abstract}

\end{frontmatter}

\section{Introduction}

Unsupervised machine learning techniques such as clustering have become widely used due to their ability to provide new insights from unlabeled datasets~\citep{Basel2015}. These algorithms aim to find global patterns (homogeneous groups) from observations based on their coherence found along all attributes~\citep{Theodoridis2009}. However, the recent \textit{big-data} phenomenon has massively increased the number and type of observations and attributes, leading to the emergence of high-dimensional and heterogeneous datasets~\citep{dhaenens2016metaheuristics,Xie2019}. For instance, in biomedicine, \textit{electronic health record} (EHR)-based research using unsupervised learning is central to fulfilling the vision of personalized medicine~\citep{Landi2020}. In these heterogeneous data problems, specific observations are usually locally correlated on a subset of attributes. In this context, biclustering is an unsupervised learning technique that simultaneously groups observations and attributes, forming \textit{biclusters}, leading to unique conceptual benefits~\citep{madeira2004biclustering,JoseGarciaJVC2023survey}: (i) can unravel local patterns, i.e., observations meaningfully correlated on a subset of attributes; (ii) provide flexibility in the definition of different types of biclusters; (iii) allow overlapping biclusters, such that one observation can belong to more than one bicluster; and (iv) help alleviate the \textit{curse of dimensionality} problem, as the space volume grows exponentially with the number of observations.

Biclustering is an active research field with application to diverse problems, such as clinical and biological~\citep{Xie2019,Vandromme2022}, gene expression and microarray data~\citep{Nepomuceno2011}, text mining~\citep{Stanislav2008}, and time series analysis~\citep{Castanho2022}. During the last 20 years, many biclustering algorithms and tools have been developed to provide insights into large high-dimensional datasets~\citep{madeira2004biclustering,Xie2019,JoseGarciaJVC2023survey}.
However, current biclustering algorithms mainly operate on numeric attributes, which poses challenges when dealing with highly heterogeneous datasets containing multiple data types, such as binary, numeric, and categorical attributes. In practice, feature transformation is a commonly used technique~\citep{Wei2015,Becue2008}; unfortunately, this early transformation step largely pre-determines the results and can cause information loss, as the relative importance of different attributes is not considered~\citep{Basel2015}. 

This paper proposes a biclustering approach capable of discovering meaningful biclusters from datasets having binary, numeric, and categorical attributes. The algorithm consists of two main stages, bicluster \textit{generation}, and \textit{model selection}. In the generation stage, a heuristic-based iterative process is used to generate multiple candidate biclusters, which may consist of different data types. Then, a fitness function is proposed based on intra-cluster variance to measure the quality of obtained biclusters with mixed-type attributes. In the second stage, we propose two strategies for selecting the most meaningful biclusters generated in the first stage. The resulting biclustering approach not only has the capability to automatically identify the number of biclusters, but can also leverage domain-specific knowledge if it is available.

The organization of this paper is as follows. Section~\ref{sec:background} presents background concepts and discusses related works. Our proposed biclustering approach is described in Section~\ref{sec:proposal}. Section~\ref{sec:setup} provides the experimental setup, benchmarks, and performance assessment. The results and main findings are discussed in Section~\ref{sec:results}. Finally, Section~\ref{sec:conclusions} concludes and presents future research directions.

\section{Background and related work}\label{sec:background}

This section presents some basic concepts, summarizes the related works, and discusses relevant literature.
\subsection{Biclusters}

A data matrix $\mathbf{X}$ is defined by $N$ observations (rows), $R = \{1,\ldots,N\}$, and $M$ attributes (columns), $C = \{1,\ldots,M\}$. A bicluster $B=\left(I,J\right)$ is a subset of rows $I\subseteq R$ and columns $J\subseteq C$ of the original matrix, and $b_{ij}$ represents a value in the bicluster corresponding to the $i$-th row and the $j$-th column. 

There exist several definitions of biclusters, each specifically tailored to different applications and data characteristics~\citep{madeira2004biclustering,CastanhoAM24}. We can identify four types of biclusters based on their assumed correlations between the bicluster values:

\begin{enumerate}

\item \textbf{Constant biclusters}: an \textit{overall-constant} bicluster has all the values equal to a constant $\pi$, i.e. $b_{ij}=\pi$; \textit{constant-rows} has all values equal per row, i.e. $b_{ij}=\alpha_{i}$; and \textit{constant-columns} has all values equal per column, i.e. $b_{ij}=\lambda_{j}$.

\item \textbf{Coherent biclusters}: an \textit{additive coherent} or \textit{shifting} bicluster is defined as $b_{ij}=\pi+\alpha_{i}+\lambda_{j}$, and a \textit{multiplicative coherent} or \textit{scaling} bicluster is defined as $b_{ij}=\pi\times\alpha_{i}\times\lambda_{j}$.

\item \textbf{Order preserving biclusters}: the rows or columns in the bicluster represent general trends in data (such as up-down-up) rather than explaining well-defined values. 

\item \textbf{Composed biclusters}: is a bicluster combining different types of biclusters. For instance, a bicluster with two sub-biclusters: an order preserving bicluster with numeric attributes and one constant-columns bicluster with categorical attributes.

\end{enumerate}

In the previous definitions, $\alpha_{i}\left(1\leq i\leq\left|I\right|\right)$ and $\lambda_{j}\left(1\leq j\leq\left|J\right|\right)$ refers to constant values used in the different bicluster definitions.

\subsection{Biclustering algorithms} \label{sub:bic-algorithms}

In real-world problems, biclustering algorithms are used to identify multiple biclusters, generating a \textit{biclustering solution}, denoted as $\mathbb{B}=\{B_1,\ldots, B_q\}$, where $q$ represents the number of biclusters. The relationships between the biclusters covering the input data matrix are determined based on two criteria: \textit{exclusivity} and \textit{exhaustivity}~\citep{madeira2004biclustering,JoseGarciaJVC2023survey}. Exclusivity indicates that a row or column belongs only to one of the $q$ bicluster in the biclustering solution. In contrast, the exhaustivity criterion specifies that every row and column belongs to at least one bicluster. Exhaustivity refers to covering the input matrix, while exclusivity is associated with biclustering all rows and columns in $\mathbf{X}$. Therefore, the bicluster types to be discovered will depend on both the problem domain and the data types involved~\citep{Noronha2022}.

Biclustering data analysis has gained much interest since the seminal work of Cheng and Church to investigate gene expression data~\citep{Cheng2000biclustering}. Nowadays, many biclustering algorithms exist in the literature, particularly for biological data such as gene expression~\citep{madeira2004biclustering,Pontes2015,JoseGarciaJVC2023survey,JoseGarciaJVC2023inbook}. In the more general case, these biclustering algorithms can be categorized into \emph{heuristic-based} and \emph{metaheuristic-based} approaches. The reader is referred to the following surveys for a detailed classification of biclustering algorithms and their applications~\citep{madeira2004biclustering,CastanhoAM24,JoseGarciaJVC2023survey}.

\subsection{Biclustering algorithm for mixed-type data}

We discuss existing biclustering algorithms from two crucial design aspects: their capacity to address multiple heterogeneous attributes and their ability to determine the number of biclusters automatically. To the best of our knowledge, the works of Vandromme et al.~\citep{Vandromme2016,Vandromme2022} and Selosse et al.~\citep{Selosse2020} are the only biclustering approaches specifically designed to extract biclusters from heterogeneous datasets. Vandromme et al.~\citep{Vandromme2016} proposed a biclustering algorithm called HBC to extract biclusters with constant values in the columns from medical data matrices, where the attributes can be numerical, symbolic, and binary. Subsequently, the HBC algorithm was extended for temporal data~\citep{Vandromme2022} (i.e., data from multiple patient visits). Although this approach is representative of the literature, it lacks a comprehensive evaluation against other traditional biclustering algorithms, and the ability of HBC to find other types of biclusters (i.e., overlapping) is not investigated. Selosse et al.~\citep{Selosse2020} presented a model-based biclustering approach for data with different types of attributes. They extended the \textit{latent block model} (LBM) from numerical data to other data types, such as categorical and binary. In addition, a distribution type was defined for each data type and modeled using a multiple LBM. A major limitation of this model is that the mixed-type attributes cannot be part of the same column cluster, as the model is based on the assumption that the attributes of the same block share the same distribution. Additionally, the multi-parameter configuration of each distribution used in the model is a complex task that varies for each data subset.

The biclustering approaches discussed above either require as input the maximum number of biclusters to be fixed in advance~\citep{Cheng2000biclustering,Li2009}, or they keep this parameter variable and generate a certain number of biclusters (usually several) without making an informed selection of the most representative ones and leave this task only to the decision maker or data expert~\citep{Vandromme2016,Vandromme2022,Selosse2020}. The former approach is only sometimes realistic, as the most suitable number of biclusters is unknown in real-world applications. The latter is more generally applicable but completely disregards any insight or domain expertise available. To address these difficulties, our approach outlined in Section~\ref{sec:proposal} can extract various types of biclusters from heterogeneous data and automatically identify the most representative.

\section{Proposed algorithm: {\HBic}}\label{sec:proposal}

This section introduces our proposed approach \HBic (\textbf{H}eterogeneous \textbf{BIC}lustering) to address datasets with mixed-type attributes such as binary, numeric, and categorical. First, \HBic generates candidate biclusters iteratively from a discretized search space. Then, \HBic selects meaningful biclusters by ranking them according to their \textit{heterogeneous intra-bicluster variance}, measured in the original heterogeneous data space.

The overall framework, principal components, complexity analysis, and source code availability of our proposed algorithm \HBic, are discussed in the following subsections.

\subsection{Overall framework}\label{sub:overall-framework}

The overall functioning of \HBic is outlined in Algorithm~\ref{alg:hbic}. In the first stage, the numeric attributes of the input data matrix are discretized; then, a heuristic search strategy generates a set of candidate biclusters, and subsequently, all repeated biclusters are removed. Then, in the second stage, the problem of determining the most representative biclusters is addressed (see Section~\ref{sub:biclusters-selection}). 

\begin{algorithm}[ht!]
\DontPrintSemicolon
\setstretch{0.9}
\KwIn{\\
\qquad $\mathbf{X}$: a data matrix with heterogeneous features\;
\qquad $\Gamma$: an array containing the data types in $\mathbf{X}$\;
\qquad $\beta$: desired number of biclusters (optional)\;
\qquad $r_{\min},c_{\min}$: minimum rows and columns in $B$;
}
\KwOut{\\
\qquad $\mathbb{B}$: a set of $\beta$ biclusters\;
}
\tcc*[l]{\textbf{STAGE I: Generation of biclusters}}
$\mathbf{M} \leftarrow$ \textsc{Discretization($\mathbf{X}$, $\Gamma$)}\; 
$\mathbb{B} \leftarrow \emptyset$ \;
\ForEach{column $\mathbf{C} \in \mathbf{M}$}{
    \ForEach{unique value $v$ in column $\mathbf{C}$}{
    $\mathbf{r} \leftarrow$ row indices from $\mathbf{C}$ with value $v$ \;
    $\mathbf{c} \leftarrow$ the index position of column $\mathbf{C}$\;
    $B \leftarrow \left(\mathbf{r},\mathbf{c}\right)$ \tcc*[r]{new bicluster}
    
    \While{$\mathbf{not}(stop)$}{
        
        \tcc*[l]{Add the best column to $B^\prime$}
        ${B}^\prime \leftarrow$ \textsc{AddColumn($B$,$\mathbf{M}$)}\;
        \If{$\langle$ \textsc{Bsize}($B^\prime$) $>$ \textsc{Bsize}($B$) $\rangle$}{$B \leftarrow B^\prime$}
        \lElse{$stop \leftarrow true$ }
    }
    \If{$\langle$ $\left|\mathbf{r}\right| \geq r_{\min}~\&~\left|\mathbf{c}\right| \geq c_{\min}$ $\rangle$}{$\mathbb{B} \leftarrow \mathbb{B}\bigcup\left\{ B\right\} $}
    
    }  
}
$\mathbb{B} \leftarrow$ \textsc{UniqueBiclusters($\mathbb{B}$)}\; 
\tcc*[l]{\textbf{STAGE II: Bicluster model selection}}
\ForEach{bicluster $B_{i} \in \mathbb{B}$}{
   $\mathbf{F}(i) \leftarrow$ \textsc{HIV}$(B_{i},\mathbf{X},\Gamma)$ \tcc*[r]{\bf Eq.~\eqref{eq:hiv}}
}
\If{$\langle$ $\beta$ provided as input $\rangle$}{$\mathbb{B} \leftarrow $ select $\beta$ best biclusters using Eq.~\eqref{eq:fitness}}
\Else{$\mathbb{B} 
\leftarrow$ \textsc{ModelSelection}($\mathbb{B}$) \tcc*[r]{\bf Section.~\ref{sub:biclusters-selection}}}

\caption{Biclustering Algorithm \HBic.}\label{alg:hbic}
\end{algorithm}

\HBic operates over a discretized search space to generate candidates during the first stage. In \HBic, the numeric attributes are discretized, simplifying the bicluster search since these attributes will have a finite number of possible values (a limited number of possible states). Since \HBic is an unsupervised learning algorithm, only unsupervised data discretization methods can be used in the \texttt{Discretization} step (line 1 in Algorithm~\ref{alg:hbic}). \HBic uses a basic, standard equal-width binning method, which divides the data domain into equal-size intervals (or bins); however, note that other discretization strategies, such as frequency-based and pattern-mining techniques~\citep{Henriques2015}, can be used at this step. Data discretization has proven to be a helpful strategy in various data mining tasks and machine learning algorithm applications~\citep{Vandromme2022}. An analysis of the impact of the discretization step is included in the \textit{Supplementary Material}.

The bicluster generation process is based on a greedy constructive heuristic (comprising lines 2-15 in Algorithm~\ref{alg:hbic}). The method starts by creating an initial bicluster with one column and all rows containing the same value for this column. Subsequently, at each step, it adds the column with the highest number of identical values for the rows of the current bicluster and deletes the rows whose values are different (see Algorithm~\ref{alg:addcolumn}). This process is repeated for each distinct value of each column. The process terminates when all columns have been processed. During this stage, the discretized matrix, $\mathbf{M}$, is used; therefore, the method operates on a finite number of possible values for each attribute. This adaptation simplifies the counting of identical values while generating candidate biclusters. At the end of the first stage, the \texttt{UniqueBiclusters} step removes the repeated biclusters since \HBic might rediscover the same bicluster when analyzing a different column of the bicluster in the data matrix.

\begin{algorithm}[tb!]
\DontPrintSemicolon
\setstretch{0.9}
\KwIn{$B(\mathbf{r},\mathbf{c}), \mathbf{M}$;}
\KwOut{$B^{\prime}$}
\ForEach{column $j \notin \mathbf{c}$}{
    $val_j \leftarrow \mathbf{C}_j(\mathbf{r})$\tcc*[l]{values in $\mathbf{C}_j$ at rows $\mathbf{r}$}  
    $H(j) \leftarrow \textrm{count}(\textrm{most frequent value in } val_j)$\; 
}
$c_{\textrm{best}} \leftarrow$ the $j$-th column with $\max(H)$\;

$\mathbf{c}^{\prime} \leftarrow \mathbf{c} \cup \{c_{\textrm{best}}\}$

$\mathbf{r}^{\prime} \leftarrow$ update $\mathbf{r}$, so that $B$ remains homogeneous\;

$B^{\prime} \leftarrow (\mathbf{r}^{\prime},\mathbf{c}^{\prime})$\;

\caption{\textsc{AddColumn($B$,$\mathbf{M}$)}} \label{alg:addcolumn}
\end{algorithm}

\subsection{Bicluster quality} \label{sub:bicluster-quality}

The initial phase of the \HBic algorithm generates multiple candidate biclusters, $\mathbb{B}$, for a given heterogeneous data problem. In the second phase, the quality of the candidate biclusters is computed in the original heterogeneous data space, enabling the selection of the most representative biclusters. In the literature, there is a range of metrics defined for numeric biclusters~\citep{Pontes2015quality}, including the well-known \textit{minimum squared error} and the \textit{average correlation function}. To measure the quality of heterogeneous biclusters, we propose the \textit{heterogeneous intra-bicluster variance} (HIV), which is comparable to the definitions used in HBC~\citep{Vandromme2016}.

Let $\mathbf{X}$ be the input data matrix with $N$ rows and $M$ columns, and let $B=\left(I,J\right)$ a bicluster expressed as a tuple of two nonempty sets. Thus, HIV evaluates the intra-bicluster homogeneity in the presence of heterogeneous attributes as follows:

\begin{equation} \label{eq:hiv}
\textrm{HIV}\left(B\right)=\textrm{ANV}\left(I,J_{\textrm{num}}\right)+\textrm{ACF}\left(I,J_{\textrm{cat}}\right) \enspace, 
\end{equation}
where $\textrm{ANV}\left(\cdot,\cdot\right)$ denotes the \textit{average numeric variance} for the numeric attributes such that $J_{\textrm{num}} \subseteq J$, and $\textrm{ACF}\left(\cdot,\cdot\right)$ represents the \textit{average categorical frequency} for the categorical and binary attributes such that $J_{\textrm{cat}} \subseteq J$. Firstly, the ANV function is given by

\begin{equation} 
\textrm{ANV}\left(I,J_{\textrm{num}}\right)=\frac{1}{\left|J_{\textrm{num}}\right|}\sum_{j\in J_{\textrm{num}}}\frac{\textrm{var}\left(b_{Ij}\right)}{\textrm{var}\left(x_{Rj}\right)} \enspace,
\end{equation}
where $\left|J_{\textrm{num}}\right|$ is the number of numeric attributes; $\textrm{var}(b_{Ij})$ denotes the variance of the numeric values for the $j$-th column in $B$; and $\textrm{var}(x_{Rj})$ denotes the variance of the numeric values for the $j$-th column in $\mathbf{X}$. Secondly, the ACF function estimates the extent to which a discrete value is correctly expressed to its biclusters. This is evaluated by counting the most frequent value in a certain discretized attribute (e.g., categorical or binary). The ACF is computed as

\begin{equation} 
\textrm{ACF}\left(I,J_{\textrm{cat}}\right)=\frac{1}{\left|J_{\textrm{cat}}\right|}\sum_{r\in J_{\textrm{cat}}}\left(1-\frac{\textrm{freq}\left(b_{Ij}\right)}{\left|I\right|}\right)  \enspace,
\end{equation}
where $\left|J_{\textrm{cat}}\right|$ indicates the number of categorical and binary attributes,  $\left|I\right|$ is number of rows, and  $\textrm{freq}\left(b_{Ij}\right)$ denotes the number of the most frequent value for the $j$-th attribute. 

The above expressions describe a quality measure for heterogeneous biclusters with numeric, binary, and categorical data types. However, if other data types exist in $\mathbf{X}$, this function will require further adaptations and analysis. HIV is an optimization function that takes values in the range $[0,\inf]$ and has to be minimized, i.e., values close to zero indicate more homogeneous biclusters.

\subsection{Model selection of biclusters} \label{sub:biclusters-selection}

The selection of the number of clusters is an important problem in unsupervised machine learning~\citep{Theodoridis2009,JoseGarciaW23}. In bicluster analysis, biclustering algorithms usually require defining this parameter \textit{a priori}~\citep{Cheng2000biclustering,madeira2004biclustering,JoseGarciaW2016survey}; however, in real-world applications, this parameter is usually unknown~\citep{JoseGarciaW2016survey,AmaralSGCMTA24}. In this work, we propose two selection strategies for automatically determining the number of biclusters:

\begin{description} [leftmargin=0cm]
\item[Distance-based selection] The fitness of a heterogeneous bicluster, $B=(I,J)$, is computed as

\begin{equation} \label{eq:fitness}
\textrm{fitness}(B) = \alpha \times \textrm{HIV}(B) + (1-\alpha) \times (1 - \textrm{Size}(B)) \enspace,
\end{equation}
where $\alpha$ is a constant weight representing a preference towards either the \textit{heterogeneous intra-bicluster variance} (HIV) or the \textit{bicluster size} (Size). HIV is determined using Eq.~\eqref{eq:hiv}, whereas the bicluster size is computed as the number of rows times the number of columns, $\left|I\right|\times\left|J\right|$. The values of $\textrm{HIV}(\cdot)$ and $\textrm{Size}(\cdot)$ should be normalized to the range $[0,1]$, and Eq.~\eqref{eq:fitness} should be minimized so that values close to zero indicate highly homogeneous biclusters with large sizes. Thus, based on the previous definitions, the distance-based selection is summarized in four main steps: (i)~the fitness of all the candidate biclusters are computed using Eq.~\eqref{eq:fitness}; (ii)~the biclusters are sorted according to their scores; (iii)~the difference between the scores of the biclusters is estimated; (iv)~the index of the most significant change (largest) of the consecutive score indicates the number of biclusters to be preserved.

\item[Pareto-based selection] Let us consider two objectives to be minimized: $f_1 =\textrm{HIV}(B)$ and $f_2 = (1-\textrm{Size}(B))$, as they were defined separately in Eq.~\eqref{eq:fitness}. Let $\mathbf{F} = (f_1, f_2)$ be the vector of objective functions. A bicluster solution $B^1 \in \mathbb{B}$ is said to be better than $B^2 \in \mathbb{B}$ (also known as $B^1$ dominates $B^2$, denoted as $B^1 \preceq B^2$) if and only if $f_i(B^1) \leq f_i(B^2)$ for all $i \in \{1,\ldots,m\}$ and  $f_i(B^1) < f_i(B^2)$ for at least one $i \in \{1,\ldots,m\}$, where $m=2$ denotes the number of objectives. A bicluster solution $B^*$ is Pareto optimal in case there does not exist a solution $B \in \mathbb{B}$ that dominates $B^*$. The set of all Pareto-optimal solutions is called the Pareto Set (PS), and $\textrm{FP} = \{\mathbf{F}(B)\mid B \in PS\}$ is called the Pareto Front~\citep{Deb2001}. In our case, all the non-dominated solutions in the Pareto Front represent the most relevant biclusters.

\end{description}

\subsection{Complexity analysis}

The main operation of the \HBic heuristic described above is \texttt{AddColumn} (see Algorithm~\ref{alg:addcolumn}), where the best column is added to a bicluster, and rows are removed to keep this bicluster perfectly homogeneous. The cost of this procedure is to search which rows of a column to be added are present in the current bicluster. This translates into $\mathcal{O}(r\times\log(r))$ worst-case complexity since looking for an element in a list has $\mathcal{O}(\log(r))$ complexity, where $r$ is the number of rows in the data matrix. At each step, the best column is added, which means all possible columns must be added to determine the optimal choice. Therefore, $\mathcal{O}(c^2)$ \texttt{AddColumn} operations are performed during the construction of each bicluster, where $c$ is the number of columns. This whole process is performed for each column and each possible value of this column as a starting point; therefore, $\mathcal{O}(c\times v)$ iterations are performed, where $v$ is the number of possible values for an attribute. This leads to a worst-case time complexity of $\mathcal{O}(r\times\log(r)\times c^3 \times v)$. However, fewer than $\mathcal{O}(c^2)$ \texttt{AddColumn} operations are performed in practice, as the method usually reaches the stopping criterion after adding a few dozen columns (even on heterogeneous real-world datasets). 
Finally, regarding both model selection strategies presented, the most time-consuming step is the computation of HIV, which is of the order of $\mathcal{O}(\beta \times r \times c)$, where $\beta$ is the number of candidate biclusters. This complexity analysis, along with experimental observations, indicates that the proposed method scales mainly with the number of columns.  The total number of rows; however, does not significantly impact the execution time.

\subsection{Source code availability}
The source code of \HBic (written in \texttt{Matlab} and \texttt{Python}) and the datasets used in our experiments are available to the research community in the following repositories: \url{https://github.com/adanjoga/hbic} and \url{https://github.com/clementchauvet/py-hbic}.

 \section{Experimental setting}\label{sec:setup}

\subsection{Datasets} \label{sub:datasets} \label{sub:chul-database}

\begin{description} [leftmargin=0cm]
\item[Heterogeneous synthetic datasets]
The following collection of heterogeneous synthetic datasets was generated using the \texttt{G-bic} tool recently proposed by Castanho et al.~\citep{CastanhoLHM23}, which is available through the repository: \url{https://github.com/jplobo1313/G-Bic/}. This collection comprises 21 data configurations organized into five categories. Each category aims to evaluate a unique aspect of datasets in medical applications. In the following descriptions, unless otherwise noted, each data set contains 1000 rows and 500 columns, with five planted biclusters, where 50\% of the attributes are numeric and 50\% categorical. The five categories of datasets are: 

\begin{itemize}

\item \textbf{Heterogeneity level} (HL). The percentage of categorical attributes in the dataset was modified as follows: HL $\in \{0,25,50,75,100\}$, where HL=0\% indicates a complete numeric dataset, and HL=100\% denotes a categorical dataset.

\item \textbf{Number of biclusters} (NB). The number of planted bicluster, $\beta^*$, was modified as: $\beta^* \in \{3,5,8,10\}$.

\item \textbf{Size of biclusters} (SB). The bicluster size $(\left| I\right| \times \left|J\right|)$ was modified as follows: SB $ \in \{(25 \times 25), (50 \times 50), (75 \times 75), (100 \times 100)$. For each configuration, three biclusters were planted: one numeric, one categorical, and one mixed.

\item \textbf{Size of datasets} (SM).
The dataset size $(N \times M)$ was modified as: $ \in \{(500 \times 250), (1000 \times 500), (1500 \times 750), (2000 \times 1000)$.

\item \textbf{Noise level} (NL). The percentage of noise level was modified both in the dataset and in the planted biclusters: NL $ \in \{5,10,15,20\}$.

\end{itemize}

This collection contains a total of 315 heterogeneous data problems, as for each of the 21 data configurations, 15 datasets were generated. The generation of the datasets and biclusters follows a uniform distribution, where the numerical variables are in the range of $[-10,+10]$. The categorical attributes follow the alphabet \{“a”, “b”, “c”, “d”, “e”, “f”, “g”, “h”, “i”, “j”\}. For each data configuration, different biclusters were planted, including numeric, categorical, and mixed. These problems are detailed in the \textit{Supplementary Material}.


\item[Heterogeneous data from systemic sclerosis patients]
The proposed biclustering algorithm is designed to find biclusters in highly heterogeneous data problems. Medical databases inherently contain clinical information from different data types. In this study, we considered the \textit{systemic sclerosis} (SSc) database of the \textit{Centre Hospitalier Universitaire de Lille} (herein referred to as \CHUL\footnote{SSc patients in the Internal Medicine Department of University Hospital of Lille, France, between October 2014 and December 2021 as part of the FHU PRECISE project (\textit{PREcision health in Complex Immune-mediated inflammatory diseaSEs}); sample collection and usage authorization, CPP 2019-A01083-54.} database). SSc is the most severe systemic autoimmune disease with the highest morbidity and mortality~\citep{Sobanski2019} and is characterized by significant heterogeneity with varying degrees of organ involvement among patients. This complicates the search for therapeutic targets, the design of clinical trials, and the management of patients. The current classification of patients is based on the extent of skin involvement but needs to be revised to describe the diversity of clinical phenotypes. This classification is based on the cutaneous extension of fibrosis and defines two groups of patients~\citep{Hinchcliff2019,Sobanski2019}: the \textit{limited cutaneous} form (lcSSc), in which the cutaneous fibrosis does not extend beyond the elbows and knees, and the \textit{diffuse cutaneous} form (dcSSc), in which the cutaneous fibrosis also affects the thorax, abdomen, or proximal parts of the limbs. A third category has been proposed for patients with paradoxically no skin involvement: \textit{sine scleroderma} (ssSSc). However, this classification does not consider the presence and severity of organ involvement~\citep{Sobanski2019}. The \CHUL database was created in 2014 and held clinical information of 550 SSc patients with regular, detailed follow-up visits recorded on a standardized case-report form. Currently, the database contains more than 1500 observations (patient visits) and nearly 400 attributes (e.g., demographic information, physical examination, laboratory exams, and medical analyses). Two experienced clinicians selected 40 relevant attributes, of which 22 are binary, 16 are numeric, and two are categorical. Next, data from the most recent visit of each patient was considered, limiting the bicluster analysis to 530 patients. As a result, the biclustering task was performed on 530 observations and 40 attributes with heterogeneous data types.

\end{description}
\subsection{Biclustering reference algorithms}

Regarding the comparison of our proposed approach with state-of-the-art algorithms, \HBic works on heterogeneous attributes (numeric, categorical, combination of both); however, most existing biclustering methods work exclusively on numerical data. For this reason, the following well-known algorithms are considered: Cheng and Church’s Algorithm (\textbf{CCA})~\citep{Cheng2000biclustering} and Large Average Submatrices (\textbf{LAS})~\citep{Shabalin2009}. These two algorithms are compared to three versions of the \HBic algorithm, namely: \textbf{HBIC}$_{\textrm{dis}}$, \textbf{HBIC}$_{\textrm{par}}$, and \textbf{HBIC}$_{\textrm{all}}$, corresponding to the model selection strategies \textit{distance-} and \textit{Pareto-based} as described in Section~\ref{sub:biclusters-selection}. We performed the following data preprocessing approach to use CCA and LAS in heterogeneous data problems. First, categorical data was encoded into binary using the one-hot encoding technique. Then, the algorithms were applied to this new numeric representation. Finally, a decoding step was performed to recover the column indices from the resulting biclusters.

\subsection{Evaluation metrics}

Section~\ref{sec:results} reports statistics computed from several independent executions performed for each biclustering method studied for every data problem considered. The ability of an algorithm to produce high-quality biclusters is assessed using three metrics. In the following definitions, let $B=\left(I,J\right)$ be a bicluster represented by a tuple of two nonempty sets such that $I\subseteq R$ and $J\subseteq C$, where $N$ is the number of rows and $M$ the columns in the input data matrix. Also, let us assume that $\mathbb{B}=\left\{ B_{i}\right\} _{i=1}^{k}$ and $\mathbb{B}^{*}=\left\{B_{i}^{*}\right\}_{i=1}^{q}$ are the obtained solution by the algorithm and the reference biclustering solution, respectively. The \textit{relevance} and \textit{recovery} metrics are defined as in~\citep{Prelic2006}:

\begin{equation} \label{eq:relevance}
\textrm{relevance}(\mathbb{B},\mathbb{B}^{*})=\left\{S_{\textrm{R}}(\mathbb{B},\mathbb{B}^{*})\times S_{\textrm{C}}(\mathbb{B},\mathbb{B}^{*})\right\}^{1/2}   \enspace, 
\end{equation} \label{eq:recovery}
\begin{equation}
\textrm{recovery}(\mathbb{B},\mathbb{B}^{*})=\textrm{relevance}(\mathbb{B}^{*},\mathbb{B}) \enspace, 
\end{equation}
where $S_{\textrm{R}}(\cdot,\cdot)$ and $S_{\textrm{C}}(\cdot,\cdot)$ are defined as follows

\begin{equation}
S_{\textrm{R}}(\mathbb{B},\mathbb{B}^{*})=\frac{1}{k}\sum_{B_{i}\in\mathbb{B}}\,\max_{B_{j}^{*}\in\mathbb{B}^{*}}\left\{ \frac{\left|I_{i}\cap I_{j}^{*}\right|}{\left|I_{i}\cup I_{j}^{*}\right|}\right\} \quad \textrm{and}
\end{equation}
\begin{equation}
S_{\textrm{C}}(\mathbb{B},\mathbb{B}^{*})=\frac{1}{k}\sum_{B_{i}\in\mathbb{B}}\,\max_{B_{j}^{*}\in\mathbb{B}^{*}}\left\{ \frac{\left|J_{i}\cap J_{j}^{*}\right|}{\left|J_{i}\cup J_{j}^{*}\right|}\right\} \enspace.
\end{equation}

The recovery metric evaluates the ability of an algorithm to discover the true biclusters, regardless of the total number of biclusters found by the algorithm. Meanwhile, the relevance metric quantifies the similarity between the obtained and the true biclusters, imposing penalties if the number of biclusters does not match between the two solutions~\citep{Prelic2006,JoseGarciaJVC2023survey}. Additionally, to further understand the performance and behavior of biclustering algorithms, we analyze a biclustering solution using the \textit{Clustering Error} metric presented by Patrikainen and Meila~\citep{PatrikainenM06} (herein renamed as \textit{biclustering error}), which is defined as:

\begin{equation}
\textrm{biclustering error}\left(\mathbb{B},\mathbb{B}^{*}\right)=\frac{\textrm{Dmax}(\mathbb{B},\mathbb{B}^{*})}{\left|\textrm{Uni}(\mathbb{B})\cup\textrm{Uni}(\mathbb{B}^{*})\right|} 
 \enspace,\label{eq:bic-error}
\end{equation}
where $\textrm{Uni}(\cdot)$ is the union set of a biclustering solution defined as

\begin{equation}
\textrm{Uni}\left(\mathbb{B}\right)=\bigcup_{B_{i}\in\mathbb{B}}I_{i}\times J_{i} \enspace.\label{eq:be-union}
\end{equation}

$\textrm{Dmax}\left(\cdot,\cdot\right)$  represents a unique relation $\left\{ x_{i},y_{i}\right\} _{i=1}^{\min\left(k,q\right)}$:

\begin{equation}
\textrm{Dmax}\left(\mathbb{B},\mathbb{B}^{*}\right)=\sum_{i=1}^{\min\left(k,q\right)}\left|I_{x_{i}}\times J_{x_{i}}\cap I_{y_{i}}^{*}\times J_{y_{i}}^{*}\right| \enspace.\label{eq:be-dmax}
\end{equation}

These three metrics analyze the pairwise co-assignment of data entities between the solution obtained and the correct partition (ground truth, which is known for the synthetic problems considered). The measures are defined in the ${[0,1]}$ range, and higher values indicate better biclustering performance. Horta and Campello~\citep{Horta2014} present a comparative study of these three and other biclustering metrics and didactically illustrate their implementation.

The non-parametric \textit{Kruskal--Wallis test} is used to investigate the statistical significance of the performance differences observed between the approaches compared. A significance level of ${\alpha=0.05}$ is considered in all the cases. The \textit{Bonferroni correction} is applied to account for multiple testing issues.

\section{Experimental results}\label{sec:results}

This section discusses the results of a series of experiments conducted to investigate the performance of the proposed biclustering algorithm for heterogeneous datasets, \HBic.

\begin{table*}[t!]
\small
\def\arraystretch{1.2}
\setlength{\tabcolsep}{0.07cm}

\caption{Performance of biclustering algorithms evaluated on 315 heterogeneous synthetic datasets, grouped into five categories. Biclustering performance is evaluated using the following metrics: \textit{relevance} (REL), \textit{recovery} (REC), and \textit{biclustering error} (BE). The best average (highest) value for each dataset and metric is shaded in gray, and the statistically best results are   highlighted in bold ($\alpha=0.05$).} \label{tab:results-synthetic}

\begin{tabular}{lccccccccccccccc}
\toprule 
\multirow{2}{*}{\textbf{Dataset}} & \multirow{2}{*}{\textbf{N x M}} & \multirow{2}{*}{\textbf{|I| x |J|}} & \multirow{2}{*}{$\beta^{*}$} &  & \multicolumn{3}{c}{\textbf{CCA}} &  & \multicolumn{3}{c}{\textbf{LAS}} &  & \multicolumn{3}{c}{\textbf{HBIC$_{\textrm{all}}$}}\tabularnewline
\cmidrule{6-8} \cmidrule{7-8} \cmidrule{8-8} \cmidrule{10-12} \cmidrule{11-12} \cmidrule{12-12} \cmidrule{14-16} \cmidrule{15-16} \cmidrule{16-16} 
 &  &  &  &  & \textbf{REL} & \textbf{REC} & \textbf{BE} &  & \textbf{REL} & \textbf{REC} & \textbf{BE} &  & \textbf{REL} & \textbf{REC} & \textbf{BE}\tabularnewline
\midrule
\midrule 
\textbf{HL0} & 1000 x 500 & 50 x 50 & 5 &  & 0.45\textpm 0.05 & 0.44\textpm 0.05 & 0.23\textpm 0.04 &  & 0.58\textpm 0.02 & 0.48\textpm 0.07 & 0.28\textpm 0.04 &  & \cellcolor{myg}0.96\textpm 0.05 & \cellcolor{myg}1.00\textpm 0.0 & \cellcolor{myg} 0.97\textpm 0.03\tabularnewline
\textbf{HL25} & 1000 x 500 & 50 x 50 & 5 &  & 0.47\textpm 0.06 & 0.47\textpm 0.06 & 0.27\textpm 0.05 &  & \cellcolor{myg}0.60\textpm 0.05 & 0.52\textpm 0.09 & 0.33\textpm 0.08 &  & 0.48\textpm 0.17 & \cellcolor{myg}1.00\textpm 0.0 & \cellcolor{myg}0.52\textpm 0.17\tabularnewline
\textbf{HL50} & 1000 x 500 & 50 x 50 & 5 &  & 0.48\textpm 0.07 & 0.48\textpm 0.07 & 0.27\textpm 0.06 &  & \cellcolor{myg}0.54\textpm 0.09 & 0.49\textpm 0.12 & 0.27\textpm 0.12 &  & \textbf{0.50\textpm 0.14} & \cellcolor{myg}1.00\textpm 0.0 & \cellcolor{myg}0.53\textpm 0.16\tabularnewline
\textbf{HL75} & 1000 x 500 & 50 x 50 & 5 &  & 0.43\textpm 0.05 & 0.42\textpm 0.06 & 0.24\textpm 0.05 &  & \cellcolor{myg}0.54\textpm 0.07 & 0.51\textpm 0.09 & 0.28\textpm 0.11 &  & 0.43\textpm 0.16 & \cellcolor{myg}1.00\textpm 0.0 & \cellcolor{myg}0.51\textpm 0.15\tabularnewline
\textbf{HL100} & 1000 x 500 & 50 x 50 & 5 &  & 0.43\textpm 0.09 & 0.43\textpm 0.09 & 0.26\textpm 0.06 &  & 0.59\textpm 0.10 & 0.58\textpm 0.10 & 0.34\textpm 0.11 &  & \cellcolor{myg}0.78\textpm 0.06 & \cellcolor{myg}1.00\textpm 0.0 & \cellcolor{myg}0.87\textpm 0.06\tabularnewline
\midrule 
\textbf{NB3} & 1000 x 500 & 50 x 50 & 3 &  & 0.50\textpm 0.09 & 0.50\textpm 0.09 & 0.31\textpm 0.07 &  & 0.48\textpm 0.09 & 0.44\textpm 0.14 & 0.24\textpm 0.13 &  & \cellcolor{myg}0.58\textpm 0.09 & \cellcolor{myg}1.00\textpm 0.0 & \cellcolor{myg}0.61\textpm 0.04\tabularnewline
\textbf{NB5} & 1000 x 500 & 50 x 50 & 5 &  & 0.48\textpm 0.07 & 0.48\textpm 0.07 & 0.27\textpm 0.07 &  & \cellcolor{myg}0.54\textpm 0.09 & 0.49\textpm 0.12 & 0.27\textpm 0.12 &  & \textbf{0.51\textpm 0.15} & \cellcolor{myg}1.00\textpm 0.0 & \cellcolor{myg}0.53\textpm 0.16\tabularnewline
\textbf{NB8} & 1000 x 500 & 50 x 50 & 8 &  & 0.41\textpm 0.04 & 0.41\textpm 0.05 & 0.22\textpm 0.03 &  & \cellcolor{myg}0.59\textpm 0.09 & 0.52\textpm 0.12 & 0.32\textpm 0.14 &  & 0.44\textpm 0.20 & \cellcolor{myg}1.00\textpm 0.0 & \cellcolor{myg}0.45\textpm 0.20\tabularnewline
\textbf{NB10} & 1000 x 500 & 50 x 50 & 10 &  & 0.39\textpm 0.04 & 0.38\textpm 0.04 & 0.20\textpm 0.04 &  & \cellcolor{myg}0.60\textpm 0.06 & 0.52\textpm 0.05 & 0.32\textpm 0.07 &  & 0.36\textpm 0.04 & \cellcolor{myg}1.00\textpm 0.0 & \cellcolor{myg}0.37\textpm 0.02\tabularnewline
\midrule 
\textbf{SB25} & 1000 x 500 & 25 x 25 & 3 &  & 0.07\textpm 0.05 & 0.07\textpm 0.04 & 0.01\textpm 0.02 &  & 0.19\textpm 0.17 & 0.19\textpm 0.17 & 0.05\textpm 0.07 &  & \cellcolor{myg}0.67\textpm 0.00 & \cellcolor{myg}1.00\textpm 0.0 & \cellcolor{myg}0.67\textpm 0.00\tabularnewline
\textbf{SB50} & 1000 x 500 & 50 x 50 & 3 &  & 0.50\textpm 0.09 & 0.50\textpm 0.09 & 0.31\textpm 0.07 &  & 0.48\textpm 0.09 & 0.44\textpm 0.14 & 0.24\textpm 0.13 &  & \cellcolor{myg}0.58\textpm 0.09 & \cellcolor{myg}1.00\textpm 0.0 & \cellcolor{myg}0.61\textpm 0.04\tabularnewline
\textbf{SB75} & 1000 x 500 & 75 x 75 & 3 &  & \cellcolor{myg}0.65\textpm 0.07 & 0.65\textpm 0.07 & \cellcolor{myg}0.46\textpm 0.07 &  & \textbf{0.62\textpm 0.07} & 0.55\textpm 0.11 & 0.36\textpm 0.10 &  & 0.28\textpm 0.03 & \cellcolor{myg}1.00\textpm 0.0 & 0.26\textpm 0.04\tabularnewline
\textbf{SB100} & 1000 x 500 & 100 x 100 & 3 &  & 0.58\textpm 0.14 & 0.58\textpm 0.13 & \textbf{0.42\textpm 0.10} &  & \cellcolor{myg}0.70\textpm 0.11 & 0.63\textpm 0.18 & \cellcolor{myg}0.46\textpm 0.19 &  & 0.21\textpm 0.01 & \cellcolor{myg}1.00\textpm 0.0 & 0.07\textpm 0.01\tabularnewline
\midrule 
\textbf{SM500} & 500 x 250 & 50 x 50 & 5 &  & 0.59\textpm 0.05 & 0.58\textpm 0.05 & \textbf{0.39\textpm 0.05} &  & \cellcolor{myg}0.66\textpm 0.09 & 0.59\textpm 0.11 & \cellcolor{myg}0.41\textpm 0.13 &  & 0.17\textpm 0.05 & \cellcolor{myg}1.00\textpm 0.0 & 0.07\textpm 0.02\tabularnewline
\textbf{SM1000} & 1000 x 500 & 50 x 50 & 5 &  & 0.48\textpm 0.07 & 0.48\textpm 0.07 & 0.27\textpm 0.06 &  & \cellcolor{myg}0.54\textpm 0.10 & 0.49\textpm 0.13 & 0.28\textpm 0.14 &  & \textbf{0.50\textpm 0.14} & \cellcolor{myg}1.00\textpm 0.0 & \cellcolor{myg}0.53\textpm 0.16\tabularnewline
\textbf{SM1500} & 1500 x 750 & 50 x 50 & 5 &  & 0.30\textpm 0.06 & 0.30\textpm 0.06 & 0.15\textpm 0.05 &  & 0.55\textpm 0.08 & 0.51\textpm 0.11 & 0.29\textpm 0.12 &  & \cellcolor{myg}0.57\textpm 0.17 & \cellcolor{myg}1.00\textpm 0.0 & \cellcolor{myg}0.56\textpm 0.17\tabularnewline
\textbf{SM2000} & 2000 x 1000 & 50 x 50 & 5 &  & 0.17\textpm 0.09 & 0.17\textpm 0.09 & 0.07\textpm 0.06 &  & 0.50\textpm 0.05 & 0.46\textpm 0.07 & 0.22\textpm 0.06 &  & \cellcolor{myg}0.56\textpm 0.17 & \cellcolor{myg}1.00\textpm 0.0 & \cellcolor{myg}0.56\textpm 0.17\tabularnewline
\midrule
\textbf{NL5} & 1000 x 500 & 50 x 50 & 3 &  & \textbf{0.50\textpm 0.08} & 0.50\textpm 0.08 & \cellcolor{myg}0.31\textpm 0.06 &  & \cellcolor{myg}0.50\textpm 0.11 & 0.46\textpm 0.15 & 0.26\textpm 0.15 &  & \cellcolor{myg}0.50\textpm 0.11 & \cellcolor{myg}0.86\textpm 0.1 & 0.09\textpm 0.03\tabularnewline
\textbf{NL10} & 1000 x 500 & 50 x 50 & 3 &  & \textbf{0.50\textpm 0.08} & 0.50\textpm 0.08 & \cellcolor{myg}0.30\textpm 0.06 &  & \cellcolor{myg}0.52\textpm 0.13 & 0.50\textpm 0.16 & 0.28\textpm 0.16 &  & 0.44\textpm 0.07 & \cellcolor{myg}0.75\textpm 0.1 & 0.06\textpm 0.02\tabularnewline
\textbf{NL15} & 1000 x 500 & 50 x 50 & 3 &  & \cellcolor{myg}0.50\textpm 0.10 & 0.50\textpm 0.10 & \cellcolor{myg}0.30\textpm 0.08 &  & \textbf{0.47\textpm 0.06} & 0.44\textpm 0.10 & 0.22\textpm 0.10 &  & 0.39\textpm 0.08 & \cellcolor{myg}0.73\textpm 0.1 & 0.05\textpm 0.02\tabularnewline
\textbf{NL20} & 1000 x 500 & 50 x 50 & 3 &  & \textbf{0.47\textpm 0.11} & 0.47\textpm 0.11 & \cellcolor{myg}0.28\textpm 0.09 &  & \cellcolor{myg}0.50\textpm 0.11 & 0.47\textpm 0.14 & 0.25\textpm 0.15 &  & 0.39\textpm 0.07 & \cellcolor{myg}0.68\textpm 0.1 & 0.04\textpm 0.01\tabularnewline
\bottomrule
\end{tabular}
\end{table*}
 

\begin{figure*}[tb!]

\centering
\includegraphics[width=1.0\textwidth]{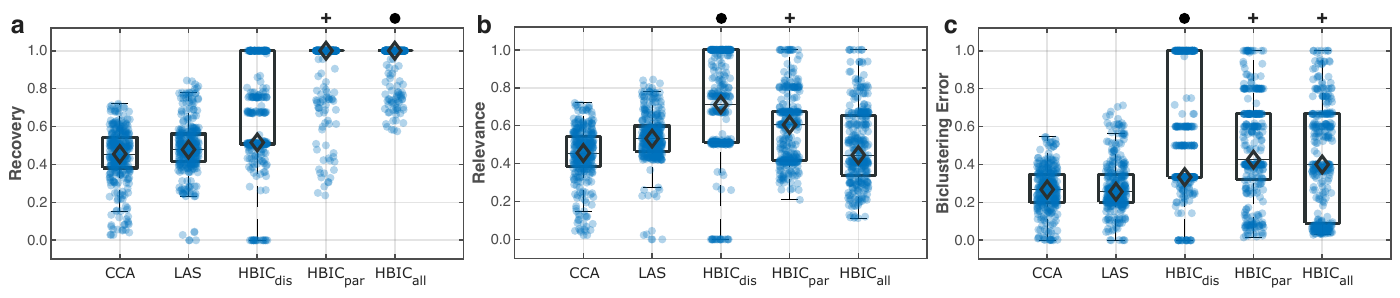}

\caption{Biclustering performance scored by the reference algorithms CCA and LAS, and \HBic versions on heterogeneous synthetic datasets in terms of the metrics (a) Recovery, (b) Relevance, and (c) Biclustering Error. Filled markets $\newmoon$ at the top of the plot, indicate the versions with the highest average value, whereas the markers $+$ denote no statistically significant differences to the best-performing algorithm.}
\label{fig:results-boxplot}

\vspace{1.0em}
\end{figure*}

\subsection{Results on heterogeneous synthetic datasets} \label{sub:results-synthetic}

This section presents a comparative analysis of the results obtained by \HBic and its performance in comparison to two reference biclustering algorithms  (see Section~\ref{sub:bic-algorithms}). The results of this comparison are summarized in Fig.~\ref{fig:results-boxplot}, while Table~\ref{tab:results-synthetic} presents detailed results for each problem configuration and category, accompanied by a statistical significance analysis. 

In the case of heterogeneous synthetic datasets, the HBIC$_{\textrm{all}}$ algorithm obtained the best performance in terms of the metric \textit{recovery}, i.e., the capacity to find the true planted biclusters, where HBIC$_{\textrm{par}}$ obtained statistically similar results. Concerning the relevance \textit{metric}, i.e., the general ability to find biclusters as good as the planted ones, the HBIC$_{\textrm{dis}}$ performed better than its counterparts. Regarding the \textit{bicluster error}, which is a compromise between the recovery and relevance metrics, the HBIC$_{\textrm{dis}}$ version also performed the best, with HBIC$_{\textrm{par}}$ and HBIC$_{\textrm{all}}$ achieving statistically similar performance. Overall, we observed that \HBic versions performed better than the CCA and LAS algorithms. The Pareto-based selection strategy performed slightly better than the distance-based.

Regarding the performance of the algorithms on the five datasets categories (see Table\ref{tab:results-synthetic}), we observe that \HBic performs remarkably well on heterogeneous datasets (heterogeneity level, HL), in data with different numbers and sizes of biclusters (NB and SM), and on datasets of different sizes (NL). However, \HBic and the benchmark algorithms present difficulties if some noise exists in the input data.

\begin{table*}
\small
\def\arraystretch{1}
\setlength{\tabcolsep}{0.165cm}

\caption{Characteristics of 10 relevant biclusters obtained by \HBic (see full list in the \textit{Supplementary Material}). These biclusters represent diverse aspects of the observed disease heterogeneity. Meanings are provided at the bottom of the table.}
\label{tab:some-bics}

\begin{tabular}{lcccccccccc}
\toprule 
\textbf{BicID} & \textbf{|I| x |J|} & \textbf{(f, m)$\dagger$} &  & \textbf{(dc, lc, ss)$\dagger$} & \textbf{mRSS} & \textbf{LVEF} & \textbf{FVC} & \textbf{DLCO} & \textbf{EUSTAR score} & \textbf{Medsger score}\tabularnewline
\midrule 
\textbf{B1} & 121x19 & (92,8) &  & (12,79,10) & 3.76\textpm 4.39 & 64.84\textpm 5.24 & 102.87\textpm 22.54 & 68.39\textpm 16.65 & 1.61\textpm 1.36 & 1.65\textpm 0.94\tabularnewline
\textbf{B7} & 309x20 & (89,11) &  & (12,74,14) & 4.03\textpm 5.12 & 64.18\textpm 5.67 & 107.70\textpm 19.49 & 73.99\textpm 21.48 & 1.55\textpm 1.24 & 1.59\textpm 0.88\tabularnewline
\textbf{B8} & 175x20 & (73,27) &  & (36,58,6) & 7.17\textpm 7.47 & 64.26\textpm 5.79 & 90.81\textpm 23.40 & 59.79\textpm 19.43 & 1.76\textpm 1.46 & 1.78\textpm 0.87\tabularnewline
\textbf{B10} & 49x19 & (80,20) &  & (22,71,6) & 7.82\textpm 7.31 & 63.45\textpm 6.74 & 83.48\textpm 21.98 & 42.38\textpm 17.12 & 2.54\textpm 1.44 & 2.12\textpm 0.93\tabularnewline
\textbf{B13} & 60x20 & (95,5) &  & (15,85,0) & 5.47\textpm 5.21 & 66.33\textpm 5.64 & 101.92\textpm 19.16 & 66.50\textpm 34.34 & 2.21\textpm 1.60 & 1.50\textpm 0.76\tabularnewline
\textbf{B15} & 96x23 & (82,18) &  & (27,55,18) & 4.34\textpm 6.17 & 64.38\textpm 5.24 & 99.42\textpm 20.98 & 70.96\textpm 18.55 & 1.50\textpm 1.38 & 1.60\textpm 0.83\tabularnewline
\textbf{B27} & 19x19 & (95,5) &  & (26,68,5) & 7.37\textpm 7.17 & 60.75\textpm 9.27 & 91.06\textpm 26.62 & 52.47\textpm 21.79 & 1.89\textpm 1.48 & 1.75\textpm 0.71\tabularnewline
\textbf{B28} & 25x20 & (92,8) &  & (16,72,12) & 4.48\textpm 5.08 & 64.83\textpm 6.11 & 91.81\textpm 23.90 & 54.15\textpm 10.90 & 1.65\textpm 1.22 & 2.00\textpm 1.05\tabularnewline
\textbf{B30} & 19x18 & (95,5) &  & (0,79,21) & 3.26\textpm 2.66 & 63.44\textpm 3.35 & 96.89\textpm 17.34 & 65.37\textpm 17.74 & 2.59\textpm 1.50 & 1.00\textpm 0.00\tabularnewline
\textbf{B38} & 246x20 & (87,13) &  & (12,70,17) & 3.35\textpm 4.72 & 64.21\textpm 5.80 & 105.38\textpm 22.20 & 73.95\textpm 23.26 & 1.54\textpm 1.30 & 1.44\textpm 0.69\tabularnewline
\bottomrule
\end{tabular}
\\

\footnotesize{Sex: m (male), f (female); cutaneous subtype: dc/lc (diffuse/limited cutaneous), ss (sine scleroderma); mRSS: mean Rodnan skin score; LVEF: left ventricular injection fraction; FVC: forced vital capacity; DLCO: diffusion lung capacity for carbon monoxide; EUSTAR: European scleroderma trials and research. Descriptive attributes that were not part of the biclustering process are indicated by the symbol ${\dagger}$.}
\end{table*}

\begin{table*}[t!]
\small
\def\arraystretch{1.0}
\setlength{\tabcolsep}{0.16cm}

\caption{Characteristics of the biclusters obtained by algorithms CCA in gray and LAS in blue. Meanings appear at the bottom of Table~\ref{tab:some-bics}.}
\label{tab:bics-cc-las}

\begin{tabular}{lcccccccccc}
\toprule 
\textbf{BicID} & \textbf{|I| x |J|} & \textbf{(f, m)$\dagger$} &  & \textbf{(dc, lc, ss)$\dagger$} & \textbf{mRSS} & \textbf{LVEF} & \textbf{FVC} & \textbf{DLCO} & \textbf{EUSTAR score} & \textbf{Medsger score}\tabularnewline
\midrule 
\rowcolor{myg}\textbf{B1} & 89x25 & (87,13) &  & (9,70,21) & 2.40\textpm 2.48 & 63.53\textpm 3.56 & 110.43\textpm 11.18 & 74.99\textpm 12.91 & 1.06\textpm 0.73 & 1.00\textpm 0.00\tabularnewline
\rowcolor{myg}\textbf{B2} & 96x21 & (90,10) &  & (8,77,15) & 2.94\textpm 2.86 & 65.82\textpm 4.46 & 106.33\textpm 13.41 & 73.15\textpm 15.04 & 1.19\textpm 1.04 & 1.33\textpm 0.71\tabularnewline
\rowcolor{myg}\textbf{B3} & 74x22 & (88,12) &  & (26,70,4) & 5.03\textpm 4.12 & 62.74\textpm 3.65 & 99.79\textpm 20.72 & 65.77\textpm 15.77 & 1.41\textpm 1.01 & 1.60\textpm 0.84\tabularnewline
\rowcolor{myg}\textbf{B4} & 72x20 & (85,15) &  & (11,74,15) & 4.66\textpm 4.80 & 65.21\textpm 5.46 & 105.50\textpm 23.22 & 76.81\textpm 20.42 & 1.46\textpm 1.29 & 1.18\textpm 0.40\tabularnewline
\rowcolor{myg}\textbf{B5} & 55x17 & (76,24) &  & (36,55,9) & 8.04\textpm 7.44 & 63.67\textpm 5.73 & 99.35\textpm 27.52 & 64.14\textpm 20.99 & 2.23\textpm 1.59 & 1.73\textpm 0.88\tabularnewline
\rowcolor{myg}\textbf{B6} & 41x17 & (78,22) &  & (29,59,12) & 4.17\textpm 5.22 & 64.42\textpm 7.36 & 101.06\textpm 22.56 & 67.65\textpm 20.13 & 1.88\textpm 1.49 & 1.56\textpm 1.01\tabularnewline
\rowcolor{myg}\textbf{B7} & 41x15 & (63,37) &  & (37,54,10) & 10.44\textpm 9.77 & 62.40\textpm 7.56 & 88.23\textpm 27.85 & 59.27\textpm 23.19 & 2.56\textpm 1.67 & 1.83\textpm 0.83\tabularnewline
\rowcolor{myg}\textbf{B8} & 39x10 & (79,21) &  & (36,59,5) & 8.99\textpm 10.52 & 64.10\textpm 7.71 & 83.00\textpm 30.24 & 52.43\textpm 19.03 & 2.40\textpm 1.56 & 1.92\textpm 0.86\tabularnewline
\rowcolor{myg}\textbf{B9} & 16x8 & (100,0) &  & (0,94,6) & 4.94\textpm 4.71 & 64.86\textpm 6.35 & 90.73\textpm 16.75 & 73.13\textpm 62.24 & 2.53\textpm 1.48 & 1.67\textpm 1.21\tabularnewline
\rowcolor{myg}\textbf{B10} & 5x14 & (100,0) &  & (20,60,20) & 2.80\textpm 2.28 & 64.00\textpm 10.4 & 93.00\textpm 16.25 & 44.00\textpm 22.02 & 1.80\textpm 1.48 & 3.00\textpm 0.00\tabularnewline
\rowcolor{myg}\textbf{B11} & 2x28 & (100,0) &  & (0,100,0) & 5.00\textpm 1.41 & 61.00\textpm 1.41 & 94.00\textpm 12.73 & 57.50\textpm 40.31 & 1.50\textpm 1.41 & --\tabularnewline
\midrule
\rowcolor{mylblue}\textbf{B$^{*}$} & 35x4 & (80,20) &  & (23,68,9) & 7.80\textpm 7.90 & 63.36\textpm 6.28 & 79.18\textpm 21.84 & 36.96\textpm 12.77 & 2.21\textpm 1.30 & 2.5\textpm 0.85\tabularnewline
\bottomrule
\end{tabular}

\end{table*}
\begin{figure}[tb!]
\centering
\includegraphics[width=0.5\textwidth]{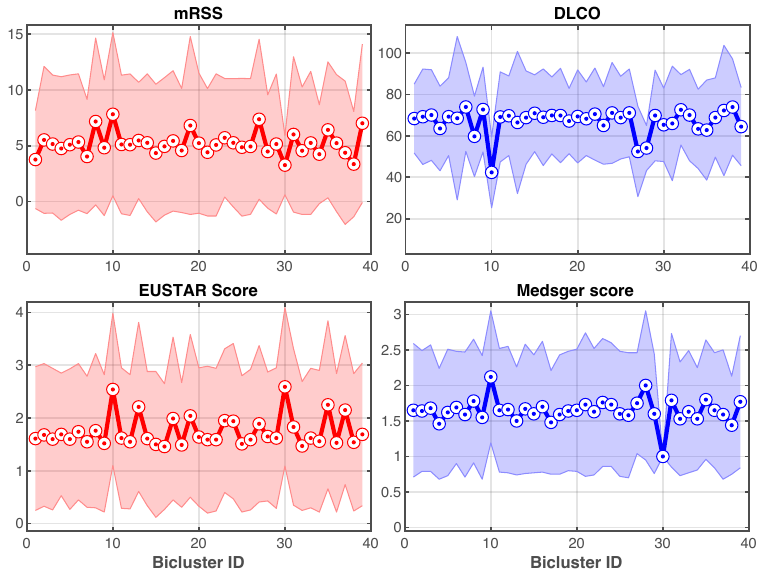}

\caption{Performance of SSc metrics (mean \textpm std) for the 39 biclusters obtained by \HBic. Meanings appear at the bottom of Table~\ref{tab:some-bics}.} \label{fig:chu-metrics}

\vspace{1.0em}
\end{figure}

\subsection{Application to systemic sclerosis} \label{sub:results-ssc}

A primary study of SSc patients suggested the existence of two (general perspective) and six (specific perspective) patient clusters~\citep{Sobanski2019, JoseGarcia2022PPSN}. It showed that there is no clear dichotomy between SSc patients with diffuse and limited cutaneous forms, and that other homogeneous groups can be found beyond skin involvement. However, these groups may not adequately represent the homogeneity of the patients because a generic clustering algorithm was used that transformed categorical variables into numerical values.

We analyzed the \texttt{CHUL} database of SSc patients (see Section~\ref{sub:chul-database}) using CCA, LAS, and \HBic biclustering algorithms. On the one hand, a total of 39 biclusters were generated using the proposed \HBic. These heterogeneous biclusters are detailed in the \textit{Supplementary Material}. The variability and differences of the obtained biclusters are illustrated in Fig.~\ref{fig:chu-metrics} in terms of four well-known metrics in SSc. A summary of 10 relevant biclusters with key clinical features is listed in Table~\ref{tab:some-bics}. On the other hand, the CCA algorithm generated 11 biclusters, while LAS found only one bicluster. It should be noted that for CCA and LAS, the categorical variables were transformed into binary variables using the one-hot-encoding technique. Details of these biclusters are reported in Table~\ref{tab:bics-cc-las}.

According to the results obtained by the three algorithms, CCA, LAS, and \HBic, it can be observed that \HBic obtained a greater diversity of biclusters of different sizes and homogeneity. It was observed that \HBic obtained an almost complete coverage of the entire clinical dataset, 95\% of the patients and 100\% of the clinical variables. While CCA obtained a patient coverage of 43\%, and its biclusters tended to cover the same region of the clinical data matrix. LAS performed very poorly on the medical data, generating only one valid bicluster (see Table~\ref{tab:bics-cc-las}, bicluster \textbf{B$^{*}$}).

In general, our preliminary results using the \HBic on the \CHUL suggest that biclustering algorithms on heterogeneous real-world datasets can provide valuable information on clinical subsetting of a disease not sufficiently explored by classical clustering methods.

\section{Conclusions} \label{sec:conclusions}

Our work proposes a biclustering approach based on a greedy heuristic for heterogeneous datasets. The resulting algorithm, \HBic, can find meaningful biclusters from heterogeneous data and automatically determines the number of biclusters when this parameter is unknown. Experiments with various datasets of different characteristics and complexities underline the robustness of the proposed approach to solving complex heterogeneous data problems. Overall, our algorithm reports a competitive performance to existing biclustering algorithms. Our approach excels in identifying meaningful biclusters from medical datasets with multiple data types.

Our analysis focused on heterogeneous data with numeric, binary, and categorical attributes. In this scenario, \HBic can retrieve constant and composed types of biclusters. However, further analysis is needed to extend this capability to other bicluster patterns and data types (i.e., longitudinal and text data). Our algorithm's design prioritized adopting straightforward yet effective techniques; nevertheless, \HBic can incorporate more sophisticated methods. For example, the selection strategies can be modified to improve the algorithm's performance. Furthermore, extending the applicability of \HBic to data with missing values is an interesting path for future investigation.


\newpage
\begin{ack}
The authors are grateful to the University of Lille, CHU Lille, and INSERM, founded by the MEL through the I-Site cluster humAIn@Lille. This work was also partly supported by the ARCHIE (\textbf{A}I for hete\textbf{R}ogeneity in \textbf{CH}ronic d\textbf{I}seas\textbf{E}s) research project.
\end{ack}

\bibliography{bib2626}

\begin{thebibliography}{35}
\providecommand{\natexlab}[1]{#1}
\providecommand{\url}[1]{\texttt{#1}}
\expandafter\ifx\csname urlstyle\endcsname\relax
  \providecommand{\doi}[1]{doi: #1}\else
  \providecommand{\doi}{doi: \begingroup \urlstyle{rm}\Url}\fi

\bibitem[Abu-Jamous et~al.(2015)Abu-Jamous, Fa, and Nandi]{Basel2015}
B.~Abu-Jamous, R.~Fa, and A.~K. Nandi.
\newblock \emph{{Integrative Cluster Analysis in Bioinformatics}}.
\newblock John Wiley \& Sons, 2015.
\newblock ISBN 978-1-118-90653-8.

\bibitem[Amaral et~al.(2024)Amaral, Soares, Gromicho, de~Carvalho, Madeira,
  Tomás, and Aidos]{AmaralSGCMTA24}
D.~M. Amaral, D.~Soares, M.~Gromicho, M.~de~Carvalho, S.~C. Madeira, P.~Tomás,
  and H.~Aidos.
\newblock {Temporal Stratification of Amyotrophic Lateral Sclerosis Patients
  Using Disease Progression Patterns}.
\newblock \emph{Nature Communication}, 15\penalty0 (1):\penalty0 5717, 2024.

\bibitem[B{\'{e}}cue{-}Bertaut and Pag{\`{e}}s(2008)]{Becue2008}
M.~B{\'{e}}cue{-}Bertaut and J.~Pag{\`{e}}s.
\newblock {Multiple Factor Analysis and Clustering of a Mixture of
  Quantitative, Categorical and Frequency data}.
\newblock \emph{Comput. Stat. Data Anal.}, 52\penalty0 (6):\penalty0
  3255--3268, 2008.

\bibitem[Busygin et~al.(2008)Busygin, Prokopyev, and Pardalos]{Stanislav2008}
S.~Busygin, O.~A. Prokopyev, and P.~M. Pardalos.
\newblock Biclustering in data mining.
\newblock \emph{Comput. Oper. Res.}, 35\penalty0 (9):\penalty0 2964--2987,
  2008.

\bibitem[Castanho et~al.(2022)Castanho, Aidos, and Madeira]{Castanho2022}
E.~Castanho, H.~Aidos, and S.~Madeira.
\newblock {Biclustering fMRI Time Series: A Comparative Study}.
\newblock \emph{{BMC} Bioinform.}, 23, 2022.

\bibitem[Castanho et~al.(2023)Castanho, Lobo, Henriques, and
  Madeira]{CastanhoLHM23}
E.~N. Castanho, J.~Lobo, R.~Henriques, and S.~C. Madeira.
\newblock {G-bic: Generating Synthetic Benchmarks for Biclustering}.
\newblock \emph{{BMC} Bioinform.}, 24\penalty0 (1):\penalty0 457, 2023.

\bibitem[Castanho et~al.(2024)Castanho, Aidos, and Madeira]{CastanhoAM24}
E.~N. Castanho, H.~Aidos, and S.~C. Madeira.
\newblock {Biclustering Data Analysis: A Comprehensive Survey}.
\newblock \emph{Briefings Bioinform.}, 25\penalty0 (4), 2024.
\newblock \doi{10.1093/BIB/BBAE342}.

\bibitem[Cheng and Church(2000)]{Cheng2000biclustering}
Y.~Cheng and G.~Church.
\newblock {Biclustering of Expression Data}.
\newblock In \emph{International Conference on Intelligent Systems for
  Molecular Biology}, volume~8, pages 93--103, 2000.

\bibitem[Deb(2001)]{Deb2001}
K.~Deb.
\newblock \emph{{Multi-objective Optimization Using Evolutionary Algorithms}}.
\newblock Wiley-Interscience series in systems and optimization. Wiley, 2001.
\newblock ISBN 978-0-471-87339-6.

\bibitem[Dhaenens and Jourdan(2016)]{dhaenens2016metaheuristics}
C.~Dhaenens and L.~Jourdan.
\newblock \emph{{Metaheuristics for Big Data}}.
\newblock John Wiley \& Sons, 2016.

\bibitem[Henriques et~al.(2015)Henriques, Antunes, and Madeira]{Henriques2015}
R.~Henriques, C.~Antunes, and S.~C. Madeira.
\newblock {A Structured View on Pattern Mining-based Biclustering}.
\newblock \emph{Pattern Recognit.}, 48\penalty0 (12):\penalty0 3941--3958,
  2015.
\newblock ISSN 0031-3203.

\bibitem[Hinchcliff and Mahoney(2019)]{Hinchcliff2019}
M.~Hinchcliff and J.~M. Mahoney.
\newblock {Towards a New Classification of Systemic Sclerosis}.
\newblock \emph{Nat Rev Rheumatol}, 15:\penalty0 456–457, 2019.

\bibitem[Horta and Campello(2014)]{Horta2014}
D.~Horta and R.~J. G.~B. Campello.
\newblock Similarity measures for comparing biclusterings.
\newblock \emph{{IEEE} {ACM} Trans. Comput. Biol. Bioinform.}, 11\penalty0 (5),
  2014.

\bibitem[{Jos{\'{e}}-Garc{\'{\i}}a} and
  {G{\'{o}}mez{-}Flores}(2016)]{JoseGarciaW2016survey}
A.~{Jos{\'{e}}-Garc{\'{\i}}a} and W.~{G{\'{o}}mez{-}Flores}.
\newblock {Automatic Clustering Using Nature-inspired Metaheuristics: A
  Survey}.
\newblock \emph{Appl. Soft Comput.}, 41:\penalty0 192--213, 2016.

\bibitem[Jos{\'{e}}-Garc{\'{\i}}a and
  G{\'{o}}mez{-}Flores(2023)]{JoseGarciaW23}
A.~Jos{\'{e}}-Garc{\'{\i}}a and W.~G{\'{o}}mez{-}Flores.
\newblock {CVIK: A Matlab-based Cluster Validity Index Toolbox for Automatic
  Data Clustering}.
\newblock \emph{SoftwareX}, 22:\penalty0 101359, 2023.
\newblock \doi{10.1016/J.SOFTX.2023.101359}.

\bibitem[{Jos{\'{e}}-Garc{\'{\i}}a} et~al.(2022){Jos{\'{e}}-Garc{\'{\i}}a},
  Jacques, Filiot, Handl, Launay, Sobanski, and Dhaenens]{JoseGarcia2022PPSN}
A.~{Jos{\'{e}}-Garc{\'{\i}}a}, J.~Jacques, A.~Filiot, J.~Handl, D.~Launay,
  V.~Sobanski, and C.~Dhaenens.
\newblock {Multi-view Clustering of Heterogeneous Health Data: Application to
  Systemic Sclerosis}.
\newblock In \emph{PPSN {XVII}: Proc. of the 17th International Conference on
  Parallel Problem Solving from Nature}, volume 13399, pages 352--367.
  Springer, 2022.

\bibitem[{Jos{\'{e}}-Garc{\'{\i}}a}
  et~al.(2023{\natexlab{a}}){Jos{\'{e}}-Garc{\'{\i}}a}, Jacques, Sobanski, and
  Dhaenens]{JoseGarciaJVC2023inbook}
A.~{Jos{\'{e}}-Garc{\'{\i}}a}, J.~Jacques, V.~Sobanski, and C.~Dhaenens.
\newblock {Biclustering Algorithms Based on Metaheuristics: A Review}.
\newblock In M.~Eddaly, B.~Jarboui, and P.~Siarry, editors,
  \emph{{Metaheuristics for Machine Learning: New Advances and Tools}}, volume
  13399, chapter~2, pages 352--367. Springer, 2023{\natexlab{a}}.
\newblock ISBN 978-981-19-3888-7.

\bibitem[{Jos{\'{e}}-Garc{\'{\i}}a}
  et~al.(2023{\natexlab{b}}){Jos{\'{e}}-Garc{\'{\i}}a}, Jacques, Sobanski, and
  Dhaenens]{JoseGarciaJVC2023survey}
A.~{Jos{\'{e}}-Garc{\'{\i}}a}, J.~Jacques, V.~Sobanski, and C.~Dhaenens.
\newblock {Metaheuristic Biclustering Algorithms: From State-of-the-art to
  Future Opportunities}.
\newblock \emph{ACM Computing Surveys}, 56\penalty0 (3), 2023{\natexlab{b}}.

\bibitem[Landi et~al.(2020)Landi, Glicksberg, Lee, Cherng, Landi, Danieletto,
  Dudley, Furlanello, and Miotto]{Landi2020}
I.~Landi, B.~S. Glicksberg, H.~Lee, S.~T. Cherng, G.~Landi, M.~Danieletto,
  J.~T. Dudley, C.~Furlanello, and R.~Miotto.
\newblock {Deep Representation Learning of Electronic Health Records to Unlock
  Patient Stratification at Scale}.
\newblock \emph{npj Digit. Medicine}, 3, 2020.

\bibitem[Li et~al.(2009)Li, Ma, Tang, Paterson, and Xu]{Li2009}
G.~Li, Q.~Ma, H.~Tang, A.~Paterson, and Y.~Xu.
\newblock {QUBIC: A Qualitative Biclustering Algorithm for Analyses of Gene
  Expression Data}.
\newblock \emph{Nucleic Acids Research}, 37, 2009.
\newblock ISSN 1362-4962.

\bibitem[Madeira and Oliveira(2004)]{madeira2004biclustering}
S.~Madeira and A.~Oliveira.
\newblock {Biclustering Algorithms for Biological Data Analysis: A Survey}.
\newblock \emph{{IEEE} {ACM} Trans. Comput. Biol. Bioinform.}, 1:\penalty0
  24--45, 2004.

\bibitem[Nepomuceno et~al.(2011)Nepomuceno, Troncoso, and
  Aguilar-Ruiz]{Nepomuceno2011}
J.~Nepomuceno, A.~Troncoso, and J.~S. Aguilar-Ruiz.
\newblock {Biclustering of Gene Expression Data by Correlation-based Scatter
  Search}.
\newblock \emph{BioData Mining}, 4\penalty0 (1), 2011.
\newblock ISSN 1756-0381.

\bibitem[Noronha et~al.(2022)Noronha, Henriques, Madeira, and
  Zárate]{Noronha2022}
M.~D. Noronha, R.~Henriques, S.~C. Madeira, and L.~E. Zárate.
\newblock {Impact of Metrics on Biclustering Solution and Quality: A Review}.
\newblock \emph{Pattern Recognit.}, 127:\penalty0 108612, 2022.
\newblock ISSN 0031-3203.

\bibitem[Patrikainen and Meila(2006)]{PatrikainenM06}
A.~Patrikainen and M.~Meila.
\newblock {Comparing Subspace Clusterings}.
\newblock \emph{{IEEE} Trans. Knowl. Data Eng.}, 18\penalty0 (7):\penalty0
  902--916, 2006.

\bibitem[Pontes et~al.(2015{\natexlab{a}})Pontes, Gir{\'{a}}ldez, and
  Aguilar{-}Ruiz]{Pontes2015}
B.~Pontes, R.~Gir{\'{a}}ldez, and J.~S. Aguilar{-}Ruiz.
\newblock {Biclustering on Expression Data: A Review}.
\newblock \emph{J. Biomed. Informatics}, 57:\penalty0 163--180,
  2015{\natexlab{a}}.

\bibitem[Pontes et~al.(2015{\natexlab{b}})Pontes, Girldez, and
  Aguilar-Ruiz]{Pontes2015quality}
B.~Pontes, R.~Girldez, and J.~S. Aguilar-Ruiz.
\newblock {Quality Measures for Gene Expression Biclusters}.
\newblock \emph{PloS One}, 10, 2015{\natexlab{b}}.

\bibitem[Preli{\'{c}} et~al.(2006)Preli{\'{c}}, Bleuler, Zimmermann, Wille,
  B{\"{u}}hlmann, Gruissem, Hennig, Thiele, and Zitzler]{Prelic2006}
A.~Preli{\'{c}}, S.~Bleuler, P.~Zimmermann, A.~Wille, P.~B{\"{u}}hlmann,
  W.~Gruissem, L.~Hennig, L.~Thiele, and E.~Zitzler.
\newblock {A Systematic Comparison and Evaluation of Biclustering Methods for
  Gene Expression Data}.
\newblock \emph{Bioinform.}, 22, 2006.
\newblock ISSN 1460-2059.

\bibitem[Selosse et~al.(2020)Selosse, Jacques, and Biernacki]{Selosse2020}
M.~Selosse, J.~Jacques, and C.~Biernacki.
\newblock {Model-based Co-clustering for Mixed Type Data}.
\newblock \emph{Comput. Stat. Data Anal.}, 144:\penalty0 106866, 2020.

\bibitem[Shabalin et~al.(2009)Shabalin, Weigman, Perou, and
  Nobel]{Shabalin2009}
A.~A. Shabalin, V.~J. Weigman, C.~M. Perou, and A.~B. Nobel.
\newblock {Finding Large Average Submatrices in High Dimensional Data}.
\newblock \emph{The Annals of Applied Statistics}, 3\penalty0 (3), 2009.
\newblock ISSN 1932-6157.

\bibitem[Sobanski et~al.(2019)Sobanski, Giovannelli, Allanore, and
  et~al.]{Sobanski2019}
V.~Sobanski, J.~Giovannelli, Y.~Allanore, and et~al.
\newblock {Phenotypes Determined by Cluster Analysis and their Survival in the
  Prospective European Scleroderma Trials and Research Cohort of Patients with
  Systemic Sclerosis}.
\newblock \emph{Arthritis \& Rheumatology}, 71, 2019.
\newblock ISSN 2326-5191.

\bibitem[Theodoridis and Koutrumbas(2009)]{Theodoridis2009}
S.~Theodoridis and K.~Koutrumbas.
\newblock \emph{{Pattern Recognition}}.
\newblock Elsevier, fourth edition, 2009.
\newblock ISBN 978-1-59749-272-0.

\bibitem[Vandromme et~al.(2016)Vandromme, Jacques, Taillard, Jourdan, and
  Dhaenens]{Vandromme2016}
M.~Vandromme, J.~Jacques, J.~Taillard, L.~Jourdan, and C.~Dhaenens.
\newblock {A Scalable Biclustering Method for Heterogeneous Medical Data}.
\newblock In \emph{{MOD}: Second International Workshop on Machine Learning,
  Optimization, and Big Data}, volume 10122, pages 70--81. Springer, 2016.

\bibitem[Vandromme et~al.(2022)Vandromme, Jacques, Taillard, Jourdan, and
  Dhaenens]{Vandromme2022}
M.~Vandromme, J.~Jacques, J.~Taillard, L.~Jourdan, and C.~Dhaenens.
\newblock {A Biclustering Method for Heterogeneous and Temporal Medical Data}.
\newblock \emph{{IEEE} Trans. Knowl. Data Eng.}, 34\penalty0 (2):\penalty0
  506--518, 2022.
\newblock ISSN 1041-4347.

\bibitem[Wei et~al.(2015)Wei, Chow, and Chan]{Wei2015}
M.~Wei, T.~Chow, and R.~Chan.
\newblock {Clustering Heterogeneous Data with K-means by Mutual
  Information-based Unsupervised Feature Transformation}.
\newblock \emph{Entropy}, 17\penalty0 (3):\penalty0 1535--1548, 2015.
\newblock ISSN 1099-4300.

\bibitem[Xie et~al.(2019)Xie, Ma, Fennell, Ma, and Zhao]{Xie2019}
J.~Xie, A.~Ma, A.~Fennell, Q.~Ma, and J.~Zhao.
\newblock {It is Time to Apply Biclustering: A Comprehensive Review of
  Biclustering Applications in Biological and Biomedical Data}.
\newblock \emph{Briefings Bioinform.}, 20\penalty0 (4):\penalty0 1449--1464,
  2019.

\end{thebibliography}
\newpage

\appendix
\section{Supplementary Material}
    
\subsection{Impact of the Discretization Strategy in \HBic}

This section aims to analyze the impact of the discretization function and its parameter $\text{nbins}$ in our proposed \HBic algorithm. The details of the \textsc{Discretization()} function are described in the main paper, and its implementation in \HBic is indicated in Algorithm 1, line 1. The parameter $\text{nbins}$ in the equal-width binning discretization helps to transform numeric attributes into discrete categories. This strategy helps to simplify the analysis of heterogeneous data; however, choosing the right intervals ($\text{nbins}$) can be challenging, as different values lead to different performances of the \HBic algorithm. In this section, we study the impact of the $\text{nbins}$ parameter in the \HBic algorithm when dealing with numerical datasets.

In our study, we considered the numerical data problems described in Section 2, and we varied the value of the parameter $\text{nbins}$ in the discretization function as $\text{nbins} = \{2,3,4,5,6,7,8,9,10,15,20\}$. Note that the $\text{nbins}$ parameter only affects datasets with numeric attributes, since the discretization step is unnecessary for categorical and binary variables. For this reason, this study of the $\text{nbins}$ parameter was performed on datasets where all attributes are numeric and represent the worst case for the \HBic algorithm, where in the best case, all attributes in the dataset are categorical or binary.

\begin{figure*}[b!]
\centering
\includegraphics[width=1\textwidth]{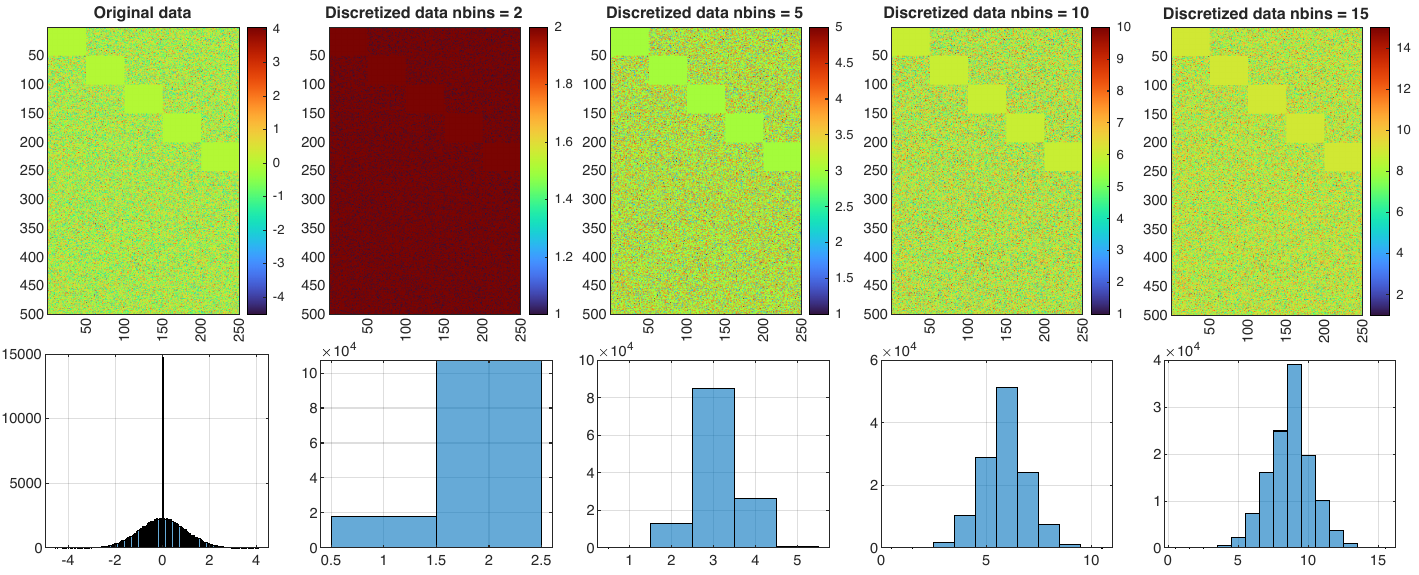}
\caption{Visualization of a numerical matrix with five biclusters and different discretization levels, $\text{nbins} = \{2,5,10,15\}$. For each data matrix, its heat map (top) and its corresponding frequency distribution (bottom) are shown. The higher the level of discretization, the closer to the original data matrix distribution.}
\label{fig:bics-discrete}
\end{figure*}

\begin{figure*}[b!]
\centering
\includegraphics[width=1\textwidth]{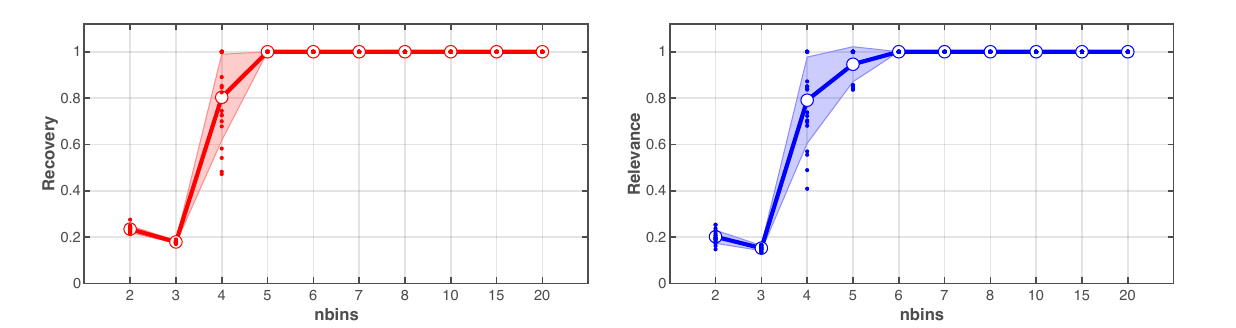}
\caption{Biclustering performance in terms of recovery (left in red) and relevance (right in blue) metrics obtained by the \HBic algorithm on the numeric datasets when varying the parameter $\text{nbins}$ in the range $\text{nbins} = \{2,3,4,5,6,7,8,9,10,15,20\}$. For both metrics, a higher value indicates better performance of the Hbic algorithm and the $\text{nbins}$ parameter. }
\label{fig:discrete-lineplot}
\end{figure*}

As a first step to visually understand the effect of the $\text{nbins}$ parameter on numeric datasets, Figure~\ref{fig:bics-discrete} exemplifies the visualization of a numerical matrix with five biclusters and different discretization levels, $\text{nbins} = \{2,5,10,15\}$. We can observe that the higher the level of discretization, the closer to the original data matrix distribution. Therefore, this indicates that a higher value of the $\text{nbins}$ parameter helps better to approximate the frequency distribution of the original data matrix; however, as the value of $\text{nbins}$ increases, the computational complexity of the \HBic algorithm will also increase.

\begin{figure*}[t!]
\centering
\includegraphics[width=1\textwidth]{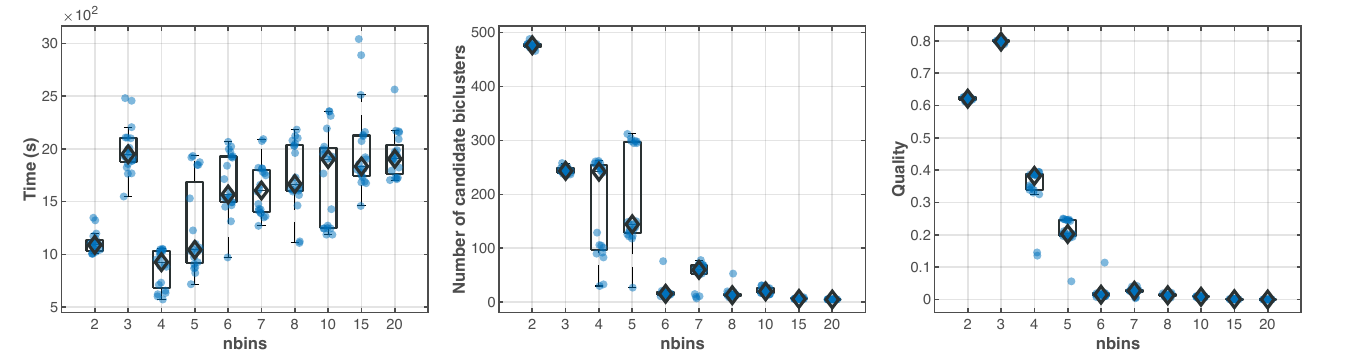}
\caption{Performance in terms of quality (left), time (center), and number of biclusters (left) obtained by the \HBic algorithm on the numerical datasets when varying the parameter $\text{nbins}$ in the range $\text{nbins} = \{2,3,4,5,6,7,8,9,10,15,20\}$. For all three metrics, a lower value indicates better performance of the \HBic algorithm and the $\text{nbins}$ parameter. }
\label{fig:discrete-boxplot}

\vspace{1.0em}
\end{figure*}

The results of the impact of the discretization strategy in the \HBic algorithm are summarized in Figures~\ref{fig:discrete-lineplot} and \ref{fig:discrete-boxplot}. Overall, from Fig.~\ref{fig:discrete-lineplot}, we can observe that as we increase the parameter $\text{nbins}$ the biclustering performance in terms of the metrics \textit{recovery} and \textit{relevance} also tends to increase. On the one hand, for the \textit{recovery} metric (the ability of the algorithm to retrieve exactly the inserted biclusters), acceptable performance close to the unity is obtained from $\text{nbins} = 5$, while for lower values the performance starts very low and increases monotonically until it becomes stable. On the other hand, regarding the \textit{relevance} metric (the ability to recover good candidate biclusters, i.e., compact with low variance), acceptable performance close to the unity is obtained from $\text{nbins} = 6$. These results suggest that the numeric datasets considered a value superior or equal to six will lead to robust biclustering performance.

\begin{table*}[tb!]
\small
\def\arraystretch{1.1}
\setlength{\tabcolsep}{0.20cm}

\caption{The 39 biclusters and their characteristics obtained by \HBic. These biclusters exhibit distinct properties representing different aspects of the heterogeneity observed in the SSc disease. Meanings appear at the bottom of the table.} \label{tab:hbic-biclustering}

\begin{tabular}{lcccccccccc}
\toprule 
\textbf{BicID} & \textbf{|I| x |J|} & \textbf{(f, m)$\dagger$} &  & \textbf{(dc, lc, ss)$\dagger$} & \textbf{mRSS} & \textbf{LVEF} & \textbf{FVC} & \textbf{DLCO} & \textbf{EUSTAR score} & \textbf{Medsger score}\tabularnewline
\midrule 
\textbf{B1} & 121x19 & (92,8) &  & (12,79,10) & 1.61\textpm 1.36 & 1.65\textpm 0.94 & 3.76\textpm 4.39 & 64.84\textpm 5.24 & 102.87\textpm 22.54 & 68.39\textpm 16.65\tabularnewline
\textbf{B2} & 368x20 & (81,19) &  & (23,65,12) & 1.68\textpm 1.35 & 1.64\textpm 0.85 & 5.52\textpm 6.58 & 64.01\textpm 5.76 & 100.98\textpm 22.44 & 69.27\textpm 23.05\tabularnewline
\textbf{B3} & 411x20 & (83,17) &  & (22,68,11) & 1.60\textpm 1.34 & 1.68\textpm 0.89 & 5.15\textpm 6.18 & 64.09\textpm 5.66 & 101.97\textpm 22.16 & 70.09\textpm 21.88\tabularnewline
\textbf{B4} & 64x21 & (86,14) &  & (14,70,16) & 1.69\textpm 1.16 & 1.46\textpm 0.78 & 4.75\textpm 6.43 & 64.86\textpm 5.70 & 99.79\textpm 25.20 & 63.60\textpm 20.43\tabularnewline
\textbf{B5} & 427x20 & (84,16) &  & (21,67,11) & 1.60\textpm 1.33 & 1.62\textpm 0.89 & 5.09\textpm 6.25 & 64.49\textpm 5.32 & 101.97\textpm 22.22 & 69.31\textpm 18.79\tabularnewline
\textbf{B6} & 49x21 & (80,20) &  & (16,71,12) & 1.74\textpm 1.29 & 1.69\textpm 0.79 & 5.33\textpm 6.11 & 62.05\textpm 7.75 & 99.32\textpm 25.54 & 68.55\textpm 39.46\tabularnewline
\textbf{B7} & 309x20 & (89,11) &  & (12,74,14) & 1.55\textpm 1.24 & 1.59\textpm 0.88 & 4.03\textpm 5.12 & 64.18\textpm 5.67 & 107.70\textpm 19.49 & 73.99\textpm 21.48\tabularnewline
\textbf{B8} & 175x20 & (73,27) &  & (36,58,6) & 1.76\textpm 1.46 & 1.78\textpm 0.87 & 7.17\textpm 7.47 & 64.26\textpm 5.79 & 90.81\textpm 23.40 & 59.79\textpm 19.43\tabularnewline
\textbf{B9} & 402x22 & (83,17) &  & (20,68,12) & 1.52\textpm 1.30 & 1.55\textpm 0.87 & 4.82\textpm 6.08 & 64.43\textpm 5.27 & 103.90\textpm 21.39 & 72.75\textpm 20.46\tabularnewline
\textbf{B10} & 49x19 & (80,20) &  & (22,71,6) & 2.54\textpm 1.44 & 2.12\textpm 0.93 & 7.82\textpm 7.31 & 63.45\textpm 6.74 & 83.48\textpm 21.98 & 42.38\textpm 17.12\tabularnewline
\textbf{B11} & 478x19 & (83,17) &  & (21,68,12) & 1.62\textpm 1.33 & 1.65\textpm 0.87 & 5.11\textpm 6.21 & 64.23\textpm 5.68 & 101.65\textpm 22.56 & 69.18\textpm 21.76\tabularnewline
\textbf{B12} & 419x20 & (82,18) &  & (21,66,13) & 1.55\textpm 1.27 & 1.66\textpm 0.89 & 5.09\textpm 6.34 & 63.87\textpm 5.61 & 101.63\textpm 23.05 & 69.74\textpm 19.21\tabularnewline
\textbf{B13} & 60x20 & (95,5) &  & (15,85,0) & 2.21\textpm 1.60 & 1.50\textpm 0.76 & 5.47\textpm 5.21 & 66.33\textpm 5.64 & 101.92\textpm 19.16 & 66.50\textpm 34.34\tabularnewline
\textbf{B14} & 373x20 & (84,16) &  & (19,71,10) & 1.61\textpm 1.27 & 1.67\textpm 0.91 & 5.26\textpm 6.18 & 64.25\textpm 5.80 & 102.72\textpm 22.60 & 68.85\textpm 22.56\tabularnewline
\textbf{B15} & 96x23 & (82,18) &  & (27,55,18) & 1.50\textpm 1.38 & 1.60\textpm 0.83 & 4.34\textpm 6.17 & 64.38\textpm 5.24 & 99.42\textpm 20.98 & 70.96\textpm 18.55\tabularnewline
\textbf{B16} & 298x20 & (79,21) &  & (22,66,12) & 1.46\textpm 1.19 & 1.70\textpm 0.92 & 4.94\textpm 6.16 & 64.26\textpm 5.73 & 99.54\textpm 23.23 & 69.03\textpm 23.34\tabularnewline
\textbf{B17} & 181x20 & (92,8) &  & (18,72,10) & 1.99\textpm 1.54 & 1.48\textpm 0.73 & 5.43\textpm 6.29 & 64.11\textpm 5.37 & 104.69\textpm 20.90 & 69.89\textpm 18.60\tabularnewline
\textbf{B18} & 390x20 & (83,17) &  & (18,69,13) & 1.49\textpm 1.18 & 1.59\textpm 0.84 & 4.57\textpm 5.56 & 64.28\textpm 5.83 & 102.51\textpm 23.26 & 69.86\textpm 22.80\tabularnewline
\textbf{B19} & 85x22 & (85,15) &  & (25,66,9) & 2.04\textpm 1.54 & 1.64\textpm 0.84 & 6.81\textpm 7.97 & 64.10\textpm 4.87 & 98.84\textpm 18.63 & 67.19\textpm 15.78\tabularnewline
\textbf{B20} & 401x20 & (83,17) &  & (21,68,11) & 1.64\textpm 1.31 & 1.65\textpm 0.86 & 5.23\textpm 6.28 & 64.20\textpm 5.79 & 101.55\textpm 22.28 & 69.41\textpm 22.46\tabularnewline
\textbf{B21} & 81x20 & (85,15) &  & (17,69,14) & 1.59\textpm 1.39 & 1.73\textpm 1.01 & 4.41\textpm 5.71 & 64.32\textpm 5.21 & 102.22\textpm 24.04 & 68.29\textpm 17.77\tabularnewline
\textbf{B22} & 392x21 & (82,18) &  & (19,69,12) & 1.59\textpm 1.35 & 1.63\textpm 0.89 & 5.07\textpm 6.37 & 64.13\textpm 5.70 & 102.51\textpm 22.47 & 70.55\textpm 22.24\tabularnewline
\textbf{B23} & 69x20 & (87,13) &  & (29,64,7) & 1.95\textpm 1.36 & 1.76\textpm 0.90 & 5.71\textpm 5.31 & 64.46\textpm 5.34 & 98.79\textpm 22.62 & 65.06\textpm 18.73\tabularnewline
\textbf{B24} & 144x20 & (90,10) &  & (21,70,9) & 1.94\textpm 1.47 & 1.73\textpm 0.87 & 5.26\textpm 5.75 & 64.53\textpm 4.72 & 100.99\textpm 19.92 & 71.12\textpm 24.35\tabularnewline
\textbf{B25} & 335x20 & (81,19) &  & (19,68,13) & 1.51\textpm 1.29 & 1.60\textpm 0.88 & 4.86\textpm 6.17 & 64.08\textpm 6.01 & 102.14\textpm 23.58 & 68.85\textpm 19.99\tabularnewline
\textbf{B26} & 426x21 & (83,17) &  & (20,68,12) & 1.59\textpm 1.33 & 1.58\textpm 0.88 & 4.93\textpm 6.08 & 64.38\textpm 5.36 & 102.85\textpm 21.96 & 71.13\textpm 21.20\tabularnewline
\textbf{B27} & 19x19 & (95,5) &  & (26,68,5) & 1.89\textpm 1.48 & 1.75\textpm 0.71 & 7.37\textpm 7.17 & 60.75\textpm 9.27 & 91.06\textpm 26.62 & 52.47\textpm 21.79\tabularnewline
\textbf{B28} & 25x20 & (92,8) &  & (16,72,12) & 1.65\textpm 1.22 & 2.00\textpm 1.05 & 4.48\textpm 5.08 & 64.83\textpm 6.11 & 91.81\textpm 23.90 & 54.15\textpm 10.90\tabularnewline
\textbf{B29} & 453x20 & (83,17) &  & (21,68,11) & 1.62\textpm 1.33 & 1.60\textpm 0.84 & 5.15\textpm 6.27 & 64.20\textpm 5.67 & 102.12\textpm 22.42 & 69.87\textpm 21.89\tabularnewline
\textbf{B30} & 19x18 & (95,5) &  & (0,79,21) & 2.59\textpm 1.50 & 1.00\textpm 0.00 & 3.26\textpm 2.66 & 63.44\textpm 3.35 & 96.89\textpm 17.34 & 65.37\textpm 17.74\tabularnewline
\textbf{B31} & 170x19 & (81,19) &  & (24,66,10) & 1.83\textpm 1.48 & 1.79\textpm 0.94 & 6.01\textpm 6.96 & 63.55\textpm 6.12 & 98.18\textpm 25.41 & 66.02\textpm 27.66\tabularnewline
\textbf{B32} & 290x23 & (84,16) &  & (18,69,12) & 1.47\textpm 1.22 & 1.53\textpm 0.80 & 4.55\textpm 5.71 & 64.58\textpm 5.28 & 104.65\textpm 20.10 & 72.63\textpm 16.99\tabularnewline
\textbf{B33} & 418x20 & (83,17) &  & (21,68,11) & 1.62\textpm 1.32 & 1.63\textpm 0.86 & 5.25\textpm 6.41 & 64.35\textpm 5.48 & 102.04\textpm 22.56 & 70.08\textpm 22.07\tabularnewline
\textbf{B34} & 62x20 & (81,19) &  & (18,68,15) & 1.56\textpm 1.34 & 1.53\textpm 0.72 & 4.24\textpm 4.43 & 63.72\textpm 6.54 & 99.10\textpm 22.70 & 63.41\textpm 19.17\tabularnewline
\textbf{B35} & 19x18 & (79,21) &  & (16,79,5) & 2.25\textpm 1.59 & 1.80\textpm 0.84 & 6.42\textpm 6.09 & 63.22\textpm 4.83 & 99.06\textpm 20.57 & 62.82\textpm 24.17\tabularnewline
\textbf{B36} & 399x20 & (82,18) &  & (22,68,11) & 1.53\textpm 1.31 & 1.65\textpm 0.81 & 5.23\textpm 6.14 & 64.21\textpm 5.73 & 101.30\textpm 22.88 & 68.88\textpm 19.14\tabularnewline
\textbf{B37} & 84x20 & (90,10) &  & (12,70,18) & 2.15\textpm 1.41 & 1.59\textpm 0.91 & 4.37\textpm 6.42 & 64.51\textpm 5.40 & 104.47\textpm 19.95 & 72.29\textpm 31.47\tabularnewline
\textbf{B38} & 246x20 & (87,13) &  & (12,70,17) & 1.54\textpm 1.30 & 1.44\textpm 0.69 & 3.35\textpm 4.72 & 64.21\textpm 5.80 & 105.38\textpm 22.20 & 73.95\textpm 23.26\tabularnewline
\textbf{B39} & 222x21 & (80,20) &  & (30,65,5) & 1.69\textpm 1.35 & 1.77\textpm 0.93 & 7.01\textpm 7.10 & 64.22\textpm 5.52 & 98.10\textpm 22.08 & 64.49\textpm 18.87\tabularnewline
\bottomrule
\end{tabular}
\\

\footnotesize{Sex: m (male), f (female); cutaneous subtype: dc/lc (diffuse/limited cutaneous), ss (sine scleroderma); mRSS: mean Rodnan skin score; LVEF: left ventricular injection fraction; FVC: forced vital capacity; DLCO: diffusion lung capacity for carbon monoxide; EUSTAR: European scleroderma trials and research. Descriptive attributes that were not part of the biclustering process are indicated by the symbol ${\dagger}$.}
\end{table*}

Furthermore, regarding the other important characteristics that describe the biclustering solutions when the $\text{nbins}$ parameter is modified, Fig~\ref{fig:discrete-boxplot} illustrates some statistics regarding the computation time (in seconds), number of candidate solutions, and the average quality of the candidate biclusters. We can observe that, as expected, the computation time increases as the value of $\text{nbins}$ increases; this is because the \HBic algorithm must explore more intervals for each numerical variable. Additionally, as  $\text{nbins}$ increases, the number of candidate biclusters generated by \HBic decreases, getting closer to the actual number of biclusters in the datasets. Finally, this decreasing behavior of the number of candidate biclusters is related to a lower average quality (i.e., a better average intra-bicluster variance). In other words, a higher value of $\text{nbins}$ tends to result in a lower number of more compact biclusters.

\newpage
\subsection{Supplementary Results}

Table~\ref{tab:hbic-biclustering} summarizes the 39 characteristics of the biclusters generated by \HBic when using the \CHUL database. This medical dataset comprises 530 observations and 40 attributes (22 binary, 16 numeric, and two categorical). The proposed algorithm obtained several biclusters with different degrees of homogeneity (where a bicluster can contain attributes of different types) and of varying sizes. Also, these biclusters cover different regions of the input data matrix. Finally, although these subclusters contain interesting properties, an expert analysis is needed in order to derive medical interpretations of their potential impact on the SSc disease. 
\end{document}